%% file: acl_latex.tex
\PassOptionsToPackage{dvipsnames}{xcolor}
\pdfoutput=1

\documentclass[11pt]{article}

\usepackage[preprint]{acl}

\usepackage{times}
\usepackage{latexsym}

\usepackage[T1]{fontenc}

\usepackage[utf8]{inputenc}

\usepackage{microtype}

\usepackage{inconsolata}

\usepackage{graphicx}
\usepackage{multirow}
\usepackage{booktabs}
\usepackage{caption}
\usepackage{lipsum}
\usepackage[dvipsnames]{xcolor}
\usepackage{xspace}
\usepackage{CJKutf8}
\usepackage[utf8]{inputenc}
\usepackage{amsfonts,amsmath,amssymb,amsthm}
\usepackage{bbm}
\usepackage{url}
\usepackage{titletoc}
\usepackage{fancyhdr}
\usepackage{enumitem}
\usepackage{subcaption}
\usepackage{nicefrac}
\usepackage{wrapfig}
\usepackage{tcolorbox}

\newcommand{\ours}{\texttt{CaughtCheating}\xspace}

\title{\ours: Is Your MLLM a Good Cheating Detective?\\Exploring the Boundary of Visual Perception and Reasoning}

\author{%
    \textbf{Ming Li}\textsuperscript{\rm 1}, 
    \textbf{Chenguang Wang}, 
    \textbf{Yijun Liang}\textsuperscript{\rm 1},
    \textbf{Xiyao Wang}\textsuperscript{\rm 1}, 
    \textbf{Yuhang Zhou}\textsuperscript{\rm 1}\\ 
    \textbf{Xiyang Wu}\textsuperscript{\rm 1}, 
    \textbf{Yuqing Zhang},
    \textbf{Ruiyi Zhang},
    \textbf{Tianyi Zhou}\\
    \textsuperscript{\rm 1}University of Maryland, College Park\\
    \texttt{minglii@umd.edu}~~~~\texttt{tianyi.david.zhou@gmail.com} \\
}

\fancypagestyle{firstpage}{
  \fancyhf{}              
  \fancyhead[C]{\small\bfseries 
  \textcolor{red}{\textbf{Disclaimer.} While our work aims to evaluate visual abilities, we acknowledge that some scenarios may involve gender-related social assumptions. The identification of suspicious clues in images does not constitute definitive evidence of infidelity or wrongdoing. We encourage open communication and mutual trust in real-life relationships.} \looseness-1
  } 
}

\begin{document}

\maketitle
\thispagestyle{firstpage}

\begin{abstract}
Recent agentic Multi-Modal Large Language Models (MLLMs) such as GPT-o3 have acheived near-ceiling scores on various existing benchmarks, motivating a demand for more challenging test tasks.  
These MLLMs have been reported to excel in a few expert-level tasks for humans, e.g., GeoGuesser, reflecting their potential as a detective 
who can notice minuscule cues in an image and weave them into coherent, situational explanations, leading to a reliable answer. 
But \textit{can they match the performance of excellent human detectives?}
To answer this question, we investigate some hard scenarios where GPT-o3 can still handle, and find a common scenario where o3's performance drops to nearly zero, which we name \textit{\ours}.
It is inspired by the social media requests that ask others to detect suspicious clues from photos shared by the poster's partner.
We conduct extensive experiments and analysis to understand why existing MLLMs lack sufficient capability to solve this kind of task. 
\ours provides a class of challenging visual perception and reasoning tasks with great value and practical usage. Success in these tasks paves the way for MLLMs to acquire human-level detective perception and reasoning capabilities.
The data and code are available at \url{https://github.com/mingliiii/CaughtCheating}.
\looseness-1

\end{abstract}

\input{section_introduction}

\input{section_boundary}

\input{section_method}

\input{section_results}

\input{section_further}

\input{section_conclusion}

\bibliography{custom}

\clearpage
\appendix
\onecolumn
\startcontents[appendix]
\setcounter{tocdepth}{2}
\printcontents[appendix]{ }{0}{\section*{Table of Contents for Appendix}}

\clearpage
\input{appendix_related_work}

\clearpage
\input{appendix_benchmark_construction}

\clearpage
\input{appendix_evaluation_metrics}


\clearpage
\input{appendix_evaluation_prompt}

\clearpage
\input{appendix_boundry_examples}

\end{document}

%% file: section_introduction.tex
\vspace{-2.2mm}
\section{Introduction}
\vspace{-2.2mm}

\begin{figure}[t]
    \centering
    \includegraphics[width=\columnwidth]{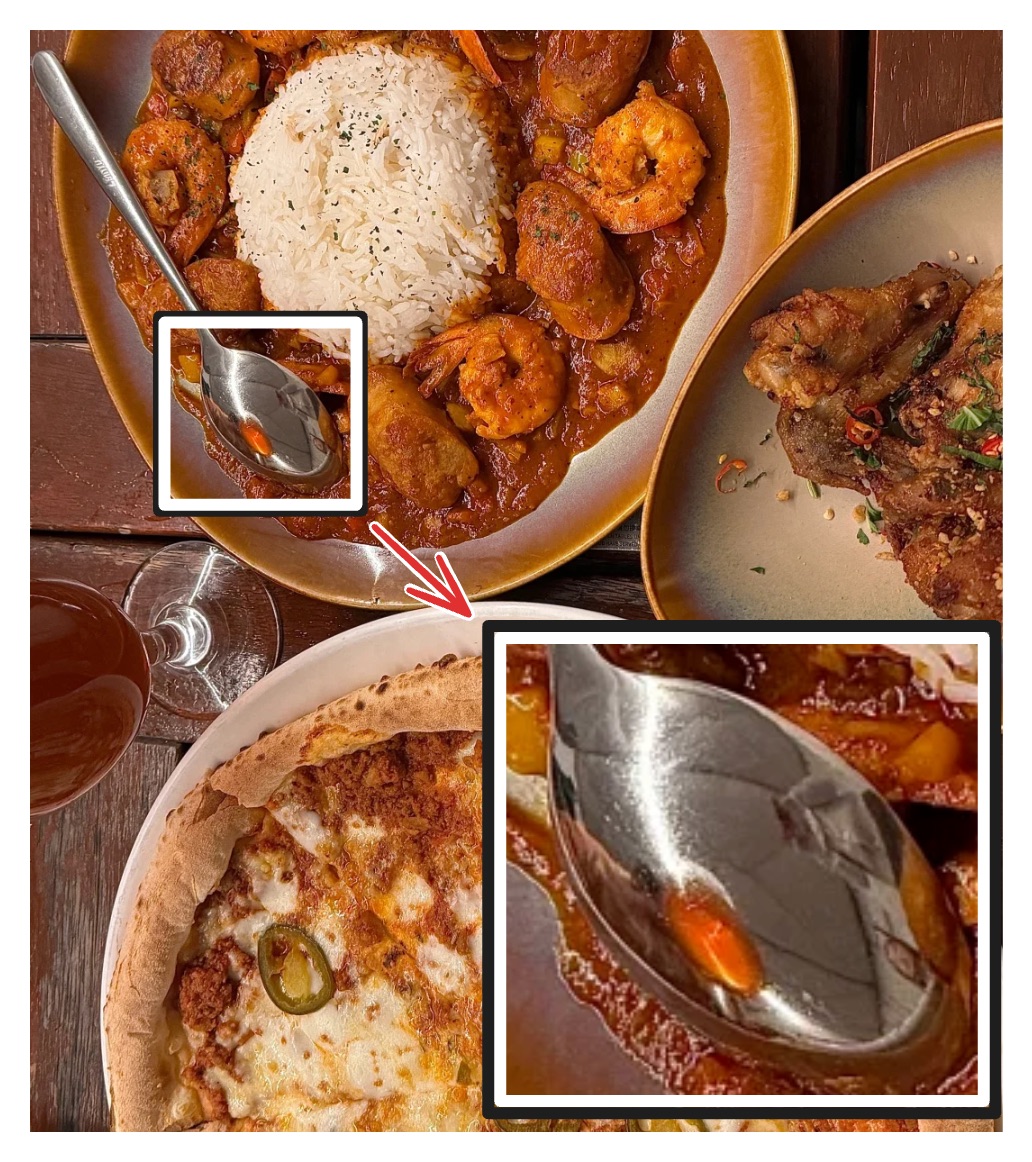} 
    \caption{
    An example from \ours. 
\textbf{Query}: ``\textit{My boyfriend said he's dining alone at the restaurant and sent me this photo. Do you notice anything suspicious in this image that contradicts his claim?}'' 
\textbf{Suspicious Clue}: ``\textit{There are other people, including a girl with long hair, visible in the spoon’s reflection.}''
In this example, most human participants, and the strong o3 are not capable of identifying this clue when not given any hints, indicating the demand of superior detective-level capabilities. \looseness-1
}
    \label{fig:intro}
    \vspace{-1.5em}
\end{figure}

Recently advanced Multi-Modal Large Language Models (MLLMs) or corresponding Agents, such as GPT-o3~\citep{openai2025o3o4systemcard} and Gemini-2.5 Pro~\citep{deepmind2025gemini25}, have demonstrated extraordinary visual perception and reasoning capabilities \citep{yue2024mmmu,zhang2024mathverse,wang2024charxiv,chen2024we, chen2024m}. 

Recent studies have demonstrated that MLLMs are even capable of addressing far more demanding challenges, e.g., GeoGuesser, estimating an image’s geographic location~\citep{luo2025doxing, huang2025vlms}. 
These kinds of tasks represent scenarios that even humans cannot accomplish easily, which require detective-level capabilities. 
These findings raise an important question: \textit{Do recent MLLMs truly acquire detective-level perception and reasoning capabilities? If so, what is the boundary of their competence?}

Motivated by Human's Last Exam~\citep{phan2025humanity}, we aim to explore and evaluate the boundary of the detective-level ability \citep{gu2023piecing, yuan2025turnaboutllm, de2025characterizing} of MLLMs on visual perception and reasoning tasks. 
We investigate a number of hard scenarios where GPT-o3 can solve the queries even though they are challenging for humans. 
Then we discover a common scenario where o3's performance drops dramatically to almost the random guess level.
This scenario is inspired by the social media requests that ask others to detect potential suspicious clues from photos shared by the poster's partner, which go against the partner's claims. 
Figure \ref{fig:intro} shows an example, in which the user query is: ``\textit{My boyfriend said he's dining alone at the restaurant and sent me this photo. Do you notice anything suspicious in this image that contradicts his claim?}'' 
This image itself seems an ordinary food-sharing image, while in the reflection of the spoon, there are other people, including a girl with long hair can be visible, which is suspicious and violates the claim of being alone. 
For this kind of task, we find that most humans, and the strong MLLMs like o3, are not able to identify the clues, indicating the superior detective-level capabilities required. \looseness-1

Thus, to explore the boundary of the visual perception and reasoning capabilities of current MLLMs \citep{johnson2017clevr, zellers2019recognition, chen2024we, chen2024m}, we collect these images and construct the \ours benchmark. 
This benchmark consists of $100$ images in total\footnote{This kind of data is intrinsically scarce. We have manually screened and verified almost all the existing related posts on public social media to construct this benchmark. }, sourced from publicly posted photographs on social media. 
The dataset is nearly evenly split into a \textit{Clued} category and a \textit{Unclued} category, and primarily features scenes from hotels and dining locations. 
Annotations for each image include a primary question about potential violation of the original claims, corresponding deterministic and non-deterministic clues, and a series of decomposed questions to analyze the visual reasoning process of MLLMs. 
\looseness-1

\ours is more challenging than the previous tasks because the targets to be identified are not directly defined in the query, and thus can not be solved by an exhaustive grid search. 
For example, when o3 tries to solve the query in Figure \ref{fig:intro}, it conducts the exhaustive grid search by focusing on one part of the figure at a time. 
However, even if it has tried focusing its attention on the area with the spoons, it still can not find this clue\footnote{o3's visual reasoning traces are presented in Appendix \ref{appendix:o3_boundry}.}. 
To theoretically analyze the difficulty dependencies between \ours and existing challenging tasks and understand the reasons behind the failures of o3, we introduce the \textit{Guided Search} theory from cognitive science~\citep{wolfe1989guided, itti2001computational, itti2002model, duncan1989visual} and the factors that guide attention in visual search.  
According to the theory, \ours has low bottom-up salience, lacks top-down feature guidance, and contains blurry scene structure and meaning. 
\looseness-1

Extensive evaluation results demonstrate that current MLLMs perform poorly on our detection-level benchmark of \ours. Notably, even the best-performing model (o3) achieved only $26.0\%$ accuracy in detecting the deterministic clues hidden in the images and $17.2\%$ IoU (the intersection over union).
Moreover, the accuracy of justifying the absence of suspicious clues (\textit{Unclued Acc}) is only $8.0\%$, resulting in the overall F1 score is only $23.9\%$.
Through investigation, we find that the current advanced MLLMs, e.g., o3 and Gemini-2.5-pro, \textbf{not only fail to identify the deterministic clues, but also tend to hallucinate and accuse everything by generating lots of so-called suspicious clues, even for innocent images}, which is not preferred.
Taken together, these results show the significance of \ours{}, which reveals that recent MLLMs still lack detective-level capabilities, and further exposes the current boundary of their visual perception and reasoning capabilities. 
Our contributions can be summarized as:

\input{figure_boundry}

\begin{itemize}[leftmargin=1em]
    \vspace{-0.8em}
    \item We systematically evaluate the limits of current MLLMs in visual perception and reasoning, analyzing how they solve various complex tasks via sophisticated reasoning strategies, and identify a common scenario where even advanced models like o3's performance drops to nearly zero.
    \vspace{-0.8em}
    \item We present \ours, the first benchmark specifically designed to assess the ability to actively search and detect subtle, context-dependent suspicious clues in real-world images. Most human annotators and state-of-the-art agentic MLLMs struggle to succeed on \ours tasks, highlighting the lack of detective-level exploration skills.
    \vspace{-0.8em}
    \item We analyze why even the most advanced agentic MLLMs fail on \ours. Inspired by the \textit{Guided Search} theory, we find that these models often lack awareness of \textit{what to search for} and \textit{how to relate observed details to the query}. Our findings offer insights into both the construction of more challenging benchmarks and the limitations of existing MLLMs.
    
\end{itemize}


%% file: figure_boundry.tex
\begin{figure*}[t]
    \centering
    \includegraphics[width=2.08\columnwidth]{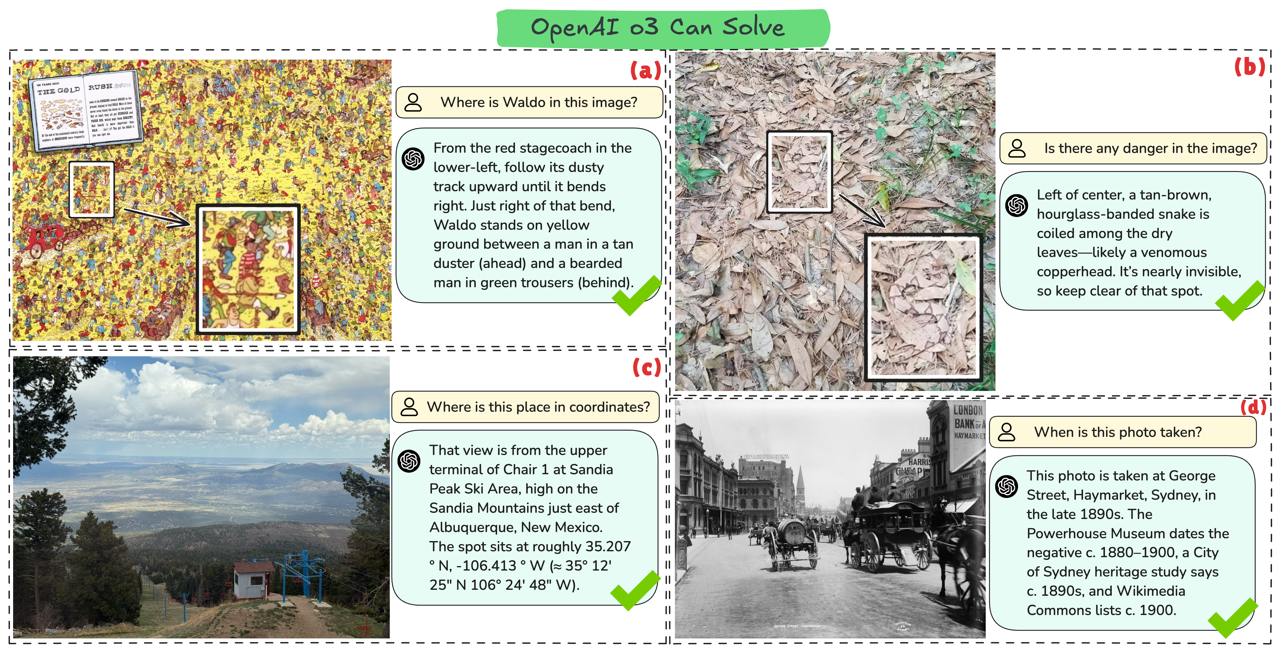}
    \caption{\textbf{Demonstration of GPT-o3’s multimodal visual-reasoning breadth.} (a) \emph{Visual search}: locating Waldo in a densely populated illustration. (b) \emph{Visual search for camouflage}: spotting a nearly invisible copperhead snake hidden among dry leaves. (c) \emph{GeoGuessr}: identifying the upper terminal of Chair 1 at New Mexico, and estimating its latitude/longitude from a single image. (d) \emph{TimeGuesser}: dating the photograph by matching architectural signage and period vehicles to museum and heritage records. These examples highlight o3's strong visual perception and reasoning capacity across various visual tasks that most humans can not accomplish. }
    \label{fig:figure_boundry}
    \vspace{-1em}
\end{figure*}

%% file: section_boundary.tex

\vspace{-2.2mm}
\section{Exploring the Boundary of Visual Perception and Reasoning}


\subsection{Reasoning Trace Analysis of o3}

As shown in the Figure \ref{fig:figure_boundry}, $4$ representative task scenarios are selected for our qualitative analysis towards the boundary of MLLM visual perception and reasoning capabilities. 
These tasks have been shown can be solved by the powerful agentic MLLM, GPT-o3, even if most of them can not be solved by individual humans\footnote{All the screenshots of o3 reasoning traces for solving these examples are provided in the Appendix \ref{appendix:o3_boundry}.}. 

When solving (a), o3 systematically sweeps the image from broad overviews to focused zooms, homing in on red-and-white horizontal stripes of the character ``Waldo''. After eliminating false matches quadrant by quadrant, it confirms Waldo's outfit and hat, then translates his pixel coordinates into an easy landmark description. 
When solving (b), the o3 methodically zooms into different areas of the leaf-litter image, from the center, lower left, and lower right, to searching for irregular shapes or patterns. Spotting rounded tan-brown coils with dark hourglass bands just left of center, then it recognize the tell-tale camouflage of a venomous pit viper (likely a copperhead). 
When solving (c), o3 compares visual clues in the photo, red chairs on blue lift towers, the wide west, facing vista over Albuquerque's grid, and the tree-line/elevation typical of Sandia Crest, with known features of Sandia Peak Ski Area. Cross-checking those details against published coordinates confirms the match.
When solving (d), o3 cross-checks catalog records for Henry King's glass-plate negatives with heritage reports that caption this very view ``c. 1890s.'' Then it matches visual clues, horse buses and a Sydney Municipal, dense telegraph wires but no electric-tram overhead, and the original Anthony Hordern's ``Palace Emporium'' sign that vanished after the $1901$ fire, to pin the scene to the year.

According to the above analysis, we find that the o3 model approaches these tasks with a \textbf{methodical, exhaustive grid search}, inspecting each region or object one by one until all plausible candidates are ruled in or out. 
However, the effectiveness of this exhaustive approach will be largely negatively affected if the target object is easily overlooked. 
Figure \ref{fig:intro} presents an example: When trying to solve the given query, o3 zooms in on the areas including pizza to confirm if slices were missing, the spoon and glass reflections to spot another diner, and the wing plate and surrounding dishes to gauge portion sizes and leftover clues. 
However, \textit{it fails to notice that there are multiple people visible in the spoon's reflection. }
Compared with other objects, the spoon is so negligible that o3 does not pay much attention to it, thus leading to the failure. 
Moreover, even occasionally, o3 coincidentally pays more attention to the spoon, it can not successfully perceive the content in the reflection. 
To conclude, we find that even though o3 is able to accomplish some complex tasks, it mainly relies on an exhaustive grid search, which indicates a lack of detective-level visual perception and reasoning capabilities.

\subsection{Guided Search Theory}

To theoretically analyze the differences between the existing visual tasks and \ours, we introduce the \textbf{\textit{Guided Search}} theory \citep{wolfe1989guided} and the corresponding factors \citep{wolfe2017five} that guide attention in visual search in the area of cognitive science. 
In its theory, searching involves directing attention to objects that might be the target. 
This process is guided to the most promising items and locations by five factors discussed in the theory: \textbf{bottom-up salience}, \textbf{top-down feature guidance}, \textbf{scene structure and meaning}, \textbf{the previous history of search}, and \textbf{the relative value of the targets and distractors}.
Through investigation on the reasoning traces of o3, we find this theory, though initially proposed in the area of cognitive science, is still applicable to the current MLLMs.
We argue that \ours is significantly more challenging than many existing visual reasoning tasks, including those depicted in Figure \ref{fig:figure_boundry}, due to the interplay of these factors. 

\textbf{Bottom-Up Salience} refers to the extent to which an item ``pops out'' from its surroundings due to its intrinsic visual properties (e.g., color, orientation, contrast). This aspect represents the easiest strategy to make visual search hard.
In examples like Figure \ref{fig:figure_boundry} (a) and (b), both the targeting objects have low bottom-up salience, making them hard to find and requiring exhaustive searches. Similarly, suspicious cues in \ours also have \textit{extremely low bottom-up salience}, like a reflection in a spoon, a partially obscured object, or a subtle item in the background, and are easily overlooked. 

\textbf{Top-Down Feature Guidance} involves using knowledge about the target's properties to guide search. Previous tasks benefit significantly from top-down guidance. For Waldo, the model searches for specific red-and-white stripes as a distinct character. For the snake, the query about ``danger'' might guide the model to look for threatening patterns.  GeoGuesser and TimeGuesser rely on identifying specific architectural styles, vegetation, or period-specific artifacts. However, this is where \ours poses a major hurdle. The ``target'', i.e., the suspicious clue, is often \textit{not a predefined object but an anomaly whose significance is context-dependent.} 
Lacking the top-down feature guidance, the model \textit{does not know what to look for} because the clue could be almost anything (an extra glass, a reflection, an out-of-place item).  
As observed, even if o3 occasionally focuses on the correct object (like the spoon), it may still fail to perceive the detail within it or infer its implication.

For \textbf{Scene Structure and Meaning}, the understanding of typical scene layouts and the relationships between objects helps guide attention to likely target locations. For previous tasks, o3 leverages scene context effectively. In GeoGuesser, it compares visual clues with known features of geographical locations.  In TimeGuesser, it matches visual clues like vehicles and signage to historical records. 
However, for \ours, the image itself might seem like an ordinary food picture or a hotel picture. Allocating the critical visual clues for the task does not merely require object recognition; it also needs to interpret subtle social cues and deviations from a presumed norm (e.g., ``dining alone''). Current MLLMs struggle with this divergent reasoning over subtle, context-dependent cues, often focusing on non-deterministic details rather than decisive evidence.

In summary, \ours is more complex due to the extremely low bottom-up salience of crucial cues, the profound lack of specific top-down feature guidance, and the need to interpret subtle social context rather than just recognizing objects or well-defined patterns.  While current agentic MLLMs can methodically search and identify objects through a process of elimination and feature matching, \ours demands a more nuanced ``detective-level'' ability to identify initially inconspicuous details and infer their significance within a specific social claim. 

%% file: section_method.tex
\section{Benchmark Construction}

\begin{figure*}[t]
    \centering
    \includegraphics[width=1\textwidth]{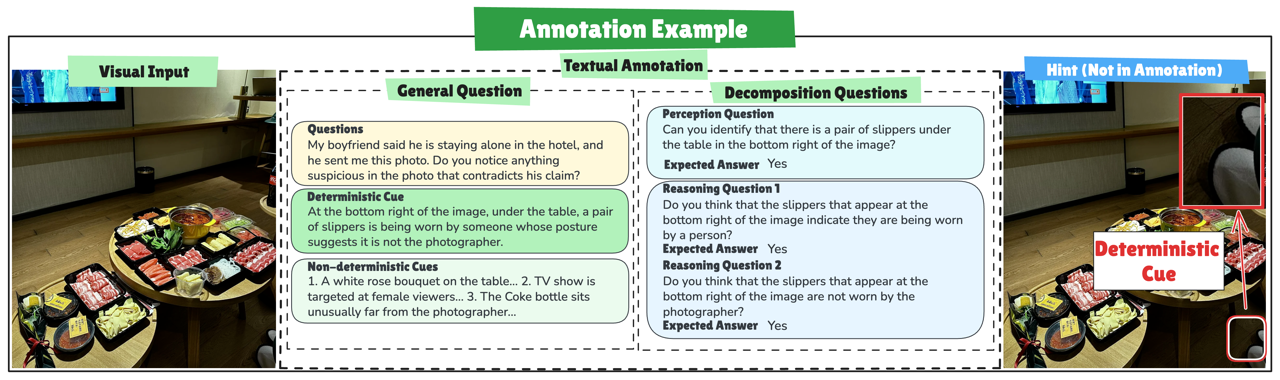}
    \caption{\textbf{An example of the annotation for the "Clued" category.} Each image is annotated with a general question assessing overall suspicion and decomposed questions focused on a deterministic clue (here, the feminine bow hair accessory). Decomposed questions include perception-based inquiries (clue identification) and reasoning-based inquiries (social implications and contradictions), all annotated with the expected answer "yes".} 
    \label{fig:exp_annot}
    \vspace{-1em}
\end{figure*}

\subsection{Image Collection}

We collect images from publicly posted photographs on social media, focusing on those posts that request others to detect potential suspicious clues that violate their partners' claims from the photos. 
We only collect images that either clearly contain or lack subtle clues related to potential violation of the claim. 
Each image is manually reviewed to ensure sufficient resolution quality for identifying such clues. 
Due to the limited availability of images with naturally occurring subtle clues, we apply minimal cropping to some images that originally show multiple people, transforming them into single-person photos while preserving subtle indicators of another person's presence. 
This approach allows us to create challenging cases where the clues are interpretable for humans but not immediately obvious. 
To ensure practical relevance, we exclude any synthetic images generated by image generation models. 
After careful selection and verification, we construct a dataset of 100 images, split into \textbf{\textit{Clued}} and \textbf{\textit{Unclued}} categories, with all personal information removed.
A detailed version of \textit{Benchmark Construction}, including the image examples, is provided in the Appendix \ref{appendix:detailed_benchmark_construction}.

\subsection{Annotation}
\label{sec:annot}

After constructing the image set, we annotate each image with a set of questions and corresponding ground-truth answers. 
A detailed annotated example is shown in Figure \ref{fig:exp_annot}. 
For images in \textit{Clued} category, we annotate each one using a question instantiated from the template: ``\textit{My [girlfriend/boyfriend] said [she/he] is [in a certain scenario] and sent me this photo. Do you notice anything suspicious in the image that contradicts [her/his] claim?}'' 
Among the potential clues, the one that deterministically shows the violation of the providing claim (a clearly identifiable, contextually inappropriate element) will be selected as the \textit{\textbf{Deterministic Clue}}, e.g., a pair of slippers is being worn by someone in Figure \ref{fig:exp_annot}. 
The remaining clues are labeled as \textit{\textbf{Non-deterministic Clues}} (weaker or more ambiguous signals), e.g., the rose bouquet, the TV shows, and the far-reached drinks.  
These non-deterministic clues might be suspicious, but apparently not enough to infer the potential claim violation. 
The reason we provide these clues is to avoid punishing models when they mention these clues. 
\looseness-1

Furthermore, we construct a series of decomposed questions designed to analyze the visual reasoning process of MLLMs, shown in the right part of Figure \ref{fig:exp_annot}. 
This series includes: 
(1) \textit{\textbf{Decomposed Perception Question}}, which assesses whether the MLLMs can identify the deterministic clue when we explicitly mention the clue and position. 
(2) \textit{\textbf{Decomposed Reasoning Question}}, which assesses whether MLLMs can understand the social implications of the clue, or whether MLLMs can imply the relation between the clue and the potential cheating.
The correct answer to each of these decomposed questions is annotated as ``yes''. 
These decomposed questions can be utilized for in-depth analysis on why MLLMs can not solve the question.

We annotate each image in the \textit{Unclued} category using the same initial question template, with ``There is no clear evidence.'' as the ground-truth answer. 


\subsection{Data Distribution}

Our dataset comprises 100 samples evenly distributed between \textit{Clued} (50) and \textit{Unclued} (50) categories. The images are captured in three common scene settings: hotels (69), dining venues (29), and karaoke bars (2). The gender distribution of photographers is balanced, with 55 male and 45 female photographers. This distribution reflects realistic patterns of photos with potential suspicious clues while maintaining a balanced representation across different categories and scenes.

\subsection{Evaluation Metrics}

We employ several evaluation metrics to comprehensively assess MLLMs' performance in detecting potential claim violations from images. 
\textbf{\textit{Clued Accuracy (Clued Acc)}} measures whether MLLMs can successfully identify the key deterministic clues in images from the \textit{Clued} category.
\textbf{\textit{Intersection over Union (Clued IoU)}} evaluates how well MLLMs identify all relevant non-deterministic clues while avoiding unrelated elements in the \textit{Clued} category.
\textbf{\textit{Unclued Accuracy (Unclued Acc)}} assesses whether MLLMs can correctly determine the absence of suspicious clues in images from the \textit{Unclued} category.
In addition to the above three metrics, we also report the accuracy of MLLMs on the decomposed questions in the analysis, including \textit{Decomposed Perception Accuracy (Dec. P Acc)}, \textit{Decomposed Reasoning Accuracy (Dec. R Acc)}, and \textit{Decomposed Overall Accuracy (Dec. Acc)} for in-depth analysis.
These metrics together provide a comprehensive evaluation framework that captures both the accuracy of clue detection and the quality of reasoning in different scenarios.

To compute these metrics, we need to parse the key points from MLLMs' open-ended responses and compare them with the ground-truth answers. Given the complexity of this task and the diversity of the responses, we recommend using human evaluators as the primary judges for the most accurate assessment. However, to enable fair and automated comparison across different models, we also develop an automatic evaluation approach using GPT-4.1 to parse and compare the model response. To validate the reliability of our automatic evaluation method, we calculate the inter-rater agreement between human evaluators and GPT-4.1 using Cohen's Kappa coefficient. The resulting kappa scores of $0.82$ for \textit{Clued Acc} and $0.943$ for \textit{Unclued Acc} demonstrate strong alignment between human and automatic evaluation, indicating the reliability of our automated assessment approach. 

Except for the provided accuracies on the two categories for an intuitive understanding of the discrepancies, we also provide the \textit{Precision}, \textit{Recall}, and \textit{F1} score for each model for a more comprehensive evaluation. 
The value of the \textit{Recall} is the same as the \textit{Clued Acc}, and the value of \textit{F1} serves as an overall metric to evaluate the performance of the model.
A detailed version of \textit{Evaluation Metrics}, including the calculation and transformation between metrics, is provided in the Appendix \ref{appendix:evaluation_metrics}.

%% file: section_results.tex
\input{table_mian_f1}
\input{table_decompsed}
\input{figure_examples}

\section{Experimental Results}

\subsection{Main Results}

The main results are shown in Table \ref{tab:main_results_f1}.
We report the accuracy and IoU on the \textit{Clued} category and accuracy on the \textit{Unclued} category. 
Models are grouped by parameter size and type (open-source vs. proprietary). 
For open-source models, we evaluate the LLaVA-OV~\citep{li2024llavaonevisioneasyvisualtask}, InternVL2~\citep{chen2024internvl}, 
InternVL2.5~\citep{chen2024internvl}, and Qwen2.5-VL~\citep{bai2025qwen25vltechnicalreport} families. 
For proprietary models, we evaluate the GPT-4o~\citep{openai2024gpt4ocard}, Gemini-2-flash~\citep{deepmind_gemini_flash}, Gemini-2.5-flash~\citep{deepmind2025gemini25}, Gemini-2.5-pro~\citep{deepmind2025gemini25}, and GPT-o3~\citep{openai2025o3o4systemcard} models. 
Human performance is also reported for reference. 

\textit{Clued Acc} and \textit{Clued IoU} represent the capability of MLLMs to identify the suspicious clues, which directly reflects the MLLMs' visual perception and reasoning abilities. 
For previous open-source models, the performance is almost negligible, as no models can reach an accuracy above 5\%, indicating their inferior capabilities on visual perception, reasoning, or even instruction following. 
As for proprietary models before the reasoning era, GPT-4o achieves 4.0\% accuracy and 1.0\% IoU, and Gemini-2-flash achieves 10.0\% accuracy and 0.0\% IoU. The performances are slightly better, indicating their better capabilities in instruction understanding and following, but still they can not reach accuracies above 10\%. 

Only for the recent strong large reasoning models, like Gemini-2.5-pro and GPT-o3, the performances can reach above 20\% accuracy and 10\% IoU, indicating their strong capabilities on visual perception and reasoning. 
But still, even the best performing model, \textit{GPT-o3}, only achieves 26.0\% accuracy and 17.2\% IoU, \textbf{indicating the current boundary of MLLMs' capabilities.} 
Considering that even human participants can only reach approximately 50\% accuracy, we believe this benchmark is challenging enough and shows the current boundary of their visual perception and reasoning capabilities. 

In the meantime, we also report the \textit{Unclued Acc} to evaluate the capability of MLLMs to not generate any suspicious clues if the image is unclued. This is also important for the real-world application, as \textbf{we do not prefer MLLMs to suspect and accuse anything if the image providers are innocent}. 
As shown in the table, most of the models reach high accuracies on this category; however, this performance is due to their inability to generate any suspicious clues. 
On the contrary, the advanced agentic models, Gemini-2.5-pro and GPT-o3, achieve low accuracies on this category, indicating their hallucination of nonexistent suspicious clues even on unclued images. 
These low accuracies reveal \textbf{their lack of strong reasoning abilities to identify if something is suspicious or not.} 
\looseness-1 

Finally, the \textit{F1} scores represent the overall performance of the model, which is the harmonic mean of the \textit{Precision} and \textit{Recall}. 
The highest \textit{F1} score is 23.9\%, which is much lower than the human performance, indicating the current boundary of MLLMs' capabilities.

\subsection{Decomposition Analysis}

To better understand why the current advanced MLLMs can not perform well in this task, we design a set of decomposed questions for each image in the \textit{Clued} category. These questions are divided into two types: perception questions, which test whether the model can accurately identify the key deterministic clue when it is explicitly mentioned, and reasoning questions, which assess whether the model can correctly infer the implications or contradictions associated.
By evaluating model performance on these questions, we can disentangle errors caused by failures in visual perception from those arising in higher-level reasoning. This fine-grained analysis helps reveal whether a model's failure is due to not seeing the clue at all, or seeing it but not understanding its significance, thus providing deeper insight into the limitations. 

As shown in Table \ref{tab:decomposed_results}, the \textit{Dec. P} is far higher than the \textit{Clued Acc}, indicating that the models can identify the key deterministic clue when it is explicitly mentioned. 
Just like how humans do during the investigation process: 
When the human participants are given the image, it's hard for them to identify the suspicious clues at the first place, e.g. the refelction in Figure \ref{fig:intro} and the femine bow hair in Figure \ref{fig:exp_annot}, but once they are explicitly mentioned or pointed out, they will admit the presense of the items. This human behavior leads to the relatively high \textit{Dec. P} for humans. 

For the \textit{Dec. R}, the performances are all relatively lower, especially for GPT-4o and GPT-o3. We find that it is because these two models refuse to answer any gender-related questions due to the safety alignment. However, \textit{even if 4o and o3 refuse to directly answer some of these questions related to genders, they might still use the gender-related information as suspicious clues.} As for the Gemini families, the \textit{Dec. R} accuracies are all relatively higher, while still lower than the \textit{Dec. P}. 

These results together indicate that current advanced MLLMs can identify the key subtle items in the image if they are explicitly mentioned. However, in \ours{}, when being asked to identify the suspicious clues without being given any hints, they tend to do an exhaustive search and generate lots of clues without really judging if the clues are suspicious or not, and at the same time, ignore the key but subtle deterministic clues. These behaviors are similar to humans and verify the hypothesis based on \textit{Guided Search} theory.

%% file: table_mian_f1.tex
\begin{table*}[t]
\centering
\resizebox{1.8\columnwidth}{!}{
\begin{tabular}{l|cc|c|ccc}
\toprule
  & \multicolumn{2}{c|}{\textbf{Clued}} & \multicolumn{1}{c|}{\textbf{Unclued}} & \multicolumn{3}{c}{\textbf{Overall}} \\
  \cmidrule(lr){2-3} \cmidrule(lr){4-4} \cmidrule(lr){5-7}
  & \textbf{Acc $\uparrow$} & \textbf{IoU $\uparrow$} & \textbf{Acc $\uparrow$} & \textbf{Precision $\uparrow$} & \textbf{Recall $\uparrow$} & \textbf{F1 $\uparrow$} \\
\midrule

\textbf{InternVL2-1B}~\citep{chen2024internvl} & 0.0 & 0.0 & 82.0 & 0.0 & 0.0 & 0.0 \\
\textbf{LLaVA-OV-1B}~\citep{li2024llavaonevisioneasyvisualtask} & 0.0 & 0.0 & 86.0 & 0.0 & 0.0 & 0.0 \\
\textbf{InternVL2.5-1B}~\citep{chen2024internvl} & 0.0 & 0.0 & 94.0 & 0.0 & 0.0 & 0.0 \\
\textbf{InternVL2-2B}~\citep{chen2024internvl} & 0.0 & 0.0 & 76.0 & 0.0 & 0.0 & 0.0 \\
\textbf{InternVL2.5-2B}~\citep{chen2024internvl} & 0.0 & 0.0 & 68.0 & 0.0 & 0.0 & 0.0 \\
\textbf{Qwen2.5-VL-3B}~\citep{bai2025qwen25vltechnicalreport} & 2.0 & 0.0 & 50.0 & 3.8 & 2.0 & 2.6 \\
\midrule
\textbf{LLaVA-v1.6-Mistral-7B}~\citep{li2024llavaonevisioneasyvisualtask}    & 0.0 & 0.0 & 82.0 & 0.0 & 0.0 & 0.0\\
\textbf{LLaVA-OV-7B}~\citep{li2024llavaonevisioneasyvisualtask}      & 2.0 & 0.0 & 52.0 & 4.0 & 2.0 & 2.7 \\
\textbf{Qwen2.5-VL-7B}~\citep{bai2025qwen25vltechnicalreport}     & 2.0 & 3.9 & 66.0 & 5.6 & 2.0 & 2.9 \\
\textbf{InternVL2-8B}~\citep{chen2024internvl}    & 0.0 & 0.0 & 76.0 & 0.0 & 0.0 & 0.0 \\
\textbf{InternVL2.5-8B}~\citep{chen2024internvl}   & 0.0 & 0.0 & 72.0 & 0.0 & 0.0 & 0.0 \\

\textbf{LLaVA-1.6-Vicuna-13B}~\citep{li2024llavaonevisioneasyvisualtask}   & 0.0 & 0.0 & 72.0 & 0.0 & 0.0 & 0.0 \\
\midrule
\textbf{InternVL2-26B}~\citep{chen2024internvl}   & 2.0 & 1.8 & 10.0 & 2.2 & 2.0 & 2.1 \\
\textbf{InternVL2.5-26B}~\citep{chen2024internvl}   & 0.0 & 0.0 & 80.0 & 0.0 & 0.0 & 0.0 \\
\textbf{InternVL2.5-38B }~\citep{chen2024internvl}   & 2.0 & 0.0 & 76.0 & 7.7 & 2.0 & 3.2 \\
\textbf{InternVL2-40B}~\citep{chen2024internvl}   & 4.0 & 0.7 & 12.0 & 4.4 & 4.0 & 4.2 \\
\textbf{InternVL2-72B}~\citep{chen2024internvl}   & 4.0 & 0.8 & 16.0 & 4.5 & 4.0 & 4.3 \\
\textbf{InternVL2.5-72B}~\citep{chen2024internvl}   & 2.0 & 0.8 & 80.0 & 9.1 & 2.0 & 3.3 \\
\textbf{LLaVA-OV-72B}~\citep{li2024llavaonevisioneasyvisualtask}    & 0.0 & 1.3 & 72.0 & 0.0 & 0.0 & 0.0 \\

\midrule

\textbf{GPT-4o}~\citep{openai2024gpt4ocard} 
& 4.0 & 1.0 & 54.0 & 8.0 & 4.0 & 5.3 \\
\textbf{Gemini-2-flash}~\citep{deepmind_gemini_flash}    & 10.0 & 0.0 & 6.0 & 9.6 & 10.0 & 9.8 \\
\textbf{Gemini-2.5-flash}~\citep{deepmind2025gemini25}  & 18.0 & 5.1 & 22.0 & 18.8 & 18.0 & 18.4 \\
\textbf{Gemini-2.5-pro}~\citep{deepmind2025gemini25}    & 20.0 & 15.1 & 22.0 & 20.4 & 20.0 & 20.2 \\
\textbf{GPT-o3}~\citep{openai2025o3o4systemcard}            & \textbf{26.0} & \textbf{17.2} & 8.0 & \textbf{22.0} & \textbf{26.0} & \textbf{23.9} \\

\midrule

\textbf{Human}    & 56.0 & / & 68.0 & 63.6 & 56.0 & 59.6 \\
\bottomrule\end{tabular}}
\caption{
The accuracies, IoU on the \textit{Clued} category, and the accuracy on the \textit{Unclued} category, and the overall precision, recall, and F1 score.
Models are grouped by parameter size and type (open-source vs. proprietary). 
\textit{Clued Acc} and \textit{IoU} represent the capability of MLLMs to identify the suspicious clues, which directly reflects the MLLMs' visual perception and reasoning abilities. 
Even the best performing model, \textit{GPT-o3}, only achieves 26.0\% accuracy and 17.2\% IoU, indicating the current boundary of MLLMs' capabilities. 
\textit{Unclued Acc} represents the capability of MLLMs to not generate any suspicious clues if the image is unclued. 
\textit{F1} score shows the overall capability of MLLMs on \ours{}, where \textit{GPT-o3}, achieves only 23.9\%.
The highest \textit{F1} score is 23.9\%, which is much lower than the human performance, indicating the current boundary of MLLMs' capabilities.
}
\label{tab:main_results_f1}
\vspace{-1em}
\end{table*} 

%% file: table_decompsed.tex
\begin{table}[t]
    \centering
    \resizebox{\columnwidth}{!}{
    \begin{tabular}{l|ccc|c}
    \midrule
      & \textbf{Dec. P} & \textbf{Dec. R} & \textbf{Dec.} & \textbf{Clued $\uparrow$} \\
    
    \midrule
    
    \textbf{GPT-4o}            & 52.0 & 12.8 & 2.0 & 4.0 \\
    \textbf{Gemini-2-flash}    & 74.0 & 69.6 & 38.0 & 10.0 \\
    \textbf{Gemini-2.5-flash}  & 72.0 & 39.2 & 20.0 & 18.0 \\
    \textbf{Gemini-2.5-pro}    & 80.0 & 52.9 & 34.0 & 20.0\\
    \textbf{GPT-o3}            & 62.0 & 24.5 & 2.0 & \textbf{26.0} \\
    
    \midrule
    \textbf{Human}    & 82.0 & 97.8 & 80.0 & 56.0 \\
    \bottomrule
    \end{tabular}
    }
    \caption{
    \textbf{Performance on decomposed questions.}
    \textit{Dec. P} and \textit{Dec. R} is the \textit{Decomposed Perception Accuracy} and \textit{Decomposed Reasoning Accuracy} of the model on the decomposed questions. 
    \textit{Dec.} is the \textit{Decomposed Accuracy} which represents the proportion of the model correctly answering all the decomposed questions. 
    }
    \label{tab:decomposed_results}
    \vspace{-1.5em}
    \end{table}

%% file: figure_examples.tex
\begin{figure*}[t]
    \centering
    \includegraphics[width=2\columnwidth]{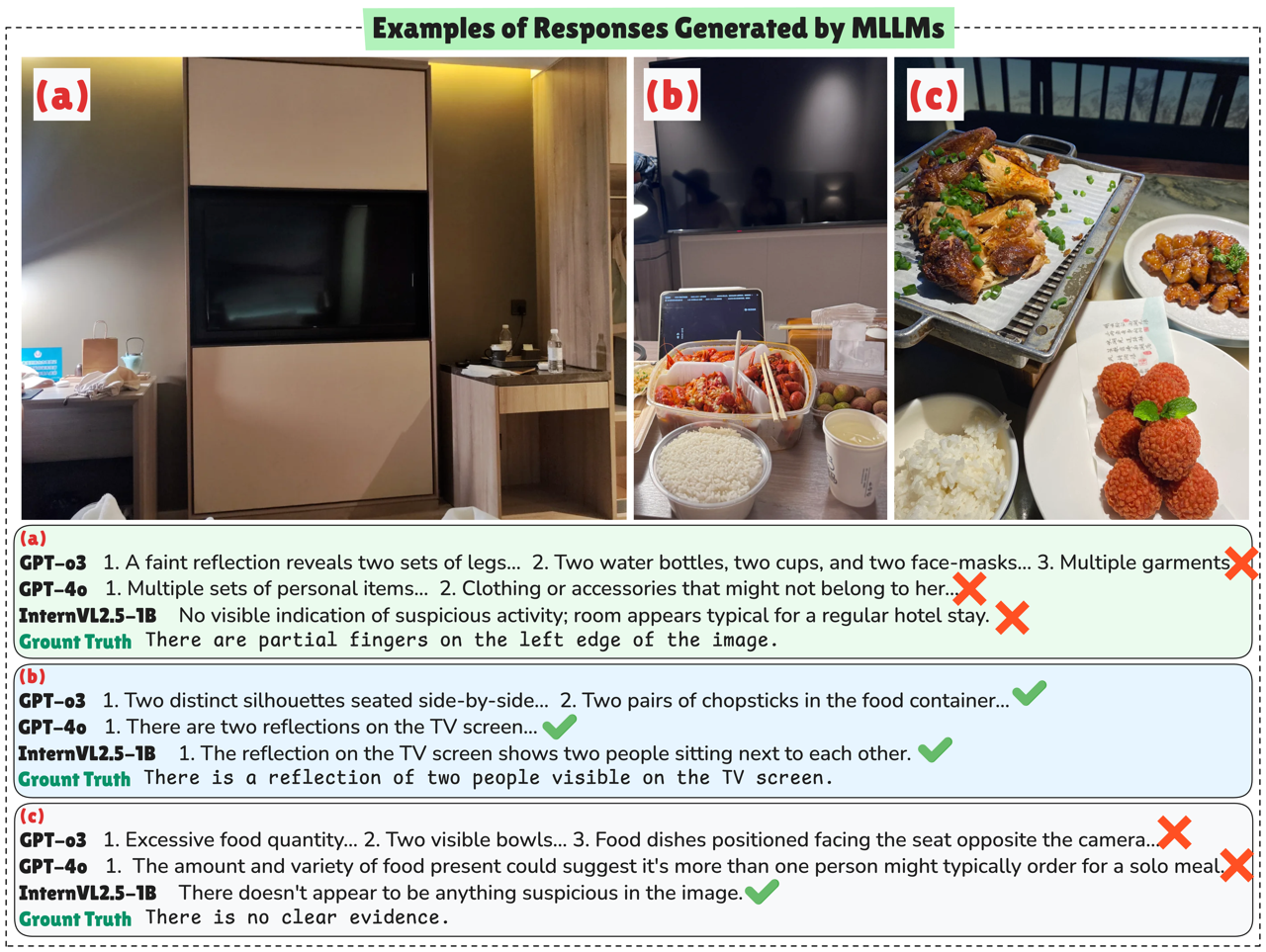}
    \caption{\textbf{Case studies of the models' performance on the \ours{} examples.} 3 representative models are selected, including GPT-o3, GPT-4o and InternVL2.5-1B, and 3 images are selected: (a) A difficult \textit{Clued} image, (b) An easy \textit{Clued} image, and (c) An \textit{Unclued} image. The models' responses are truncated for better visualization. }
    \label{fig:figure_examples}
    \vspace{-1em}
\end{figure*}

%% file: section_further.tex
\section{Case Studies}

In this section, we provide some examples to show how exactly different models perform on \ours{}, shown in Figure \ref{fig:figure_examples}.
In the figure, 3 representative models are selected, including GPT-o3, GPT-4o and InternVL2.5-1B, and 3 images are selected: (a) A difficult \textit{Clued} image, (b) An easy \textit{Clued} image, and (c) An \textit{Unclued} image.

In (a), there is an elbow, and fingers are visible at the left edge of the photo, clearly indicating the presence of another person. However, all the models fail to identify this subtle but deterministic clue and focus on the reflection of the television, even though there are no visible clues in the reflection, as another person is sitting by the table. 
What's worse, o3 and 4o keep mentioning the two bottles or cups, which are obviously provided by the hotel and can not be the suspicious clues. On the contrary, InternVL2.5-1B can not provide any clues by saying this is a normal hotel image.
In (b), the reflection in TV clearly shows there are two people on the bed, thus all the selected models can identify this clue. 
These 2 examples, (a) and (b), show that: 
(1) models are able to see through reflections, and
(2) Reflection does not always contain suspicious clues, which further verifies that \textbf{\ours{} is challenging since there are no fixed rules for the suspicious clues.}

(c) shows an \textit{Uncled} image, which is merely an ordinary food-sharing image. However, o3 still tries to generate a lot of \textit{so-called suspicious clues}, including the amount of food, the place settings, and etc. \textbf{This behaviour is not expected since we only want models to generate clues really suspecious, rather than accusing everything, which further indicates the values of \ours{}.} Similarly to the above examples, InternVL2.5-1B can not provide any clues by saying this is a normal food-sharing image, that's why it reaches the highest on the \textit{Unclued Acc}.

%% file: section_conclusion.tex
\section{Conclusion}
In this work, we present \ours, a novel benchmark designed to evaluate the capabilities of MLLMs in detecting subtle, context-dependent suspicious clues in real-world images. 
Our experiments reveal that even state-of-the-art models, such as o3, consistently fail to identify the hidden clues within these photos, suggesting the current boundary of the detective-level ability of MLLMs on visual perception and reasoning. 


\clearpage
\section*{Ethical Concern and Limitation}

Because our benchmark relies exclusively on publicly available, annotatable social-media photographs, the source pool overwhelmingly features cisgender, heterosexual couples; the scarcity of labeled images depicting LGBTQ+ or non-monogamous relationships, therefore, forced us to center this demographic. The same data constraint limited us to a handful of commonplace settings, such as hotels, restaurants, cafés, and vacation scenes, so contexts such as nightlife, workplaces, or culturally specific environments remain undersampled. Finally, the benchmark targets one complex form of visual reasoning rooted in a particular social norm, detecting suspected infidelity, without extending to the wider spectrum of complex social reasoning inferences people could draw from images. These constraints arise from limited public data, and our future benchmarks will build on more diverse, richly annotated datasets that broaden demographic coverage, scenario variety, and the range of socially grounded visual-reasoning tasks.

%% file: appendix_related_work.tex
\section{Relate Work}


\subsection{LLM reasoning}

The chain-of-thought technique~\citep{wei2022chain, kojima2022large} represents the early efforts in exploring the reasoning capabilities of large language models (LLMs)~\citep{brown2020language, chowdhery2023palm}. By explicitly generating intermediate reasoning steps, this method notably enhances performance across various reasoning tasks~\citep{patel2021nlp, cobbe2021training}. 
Moreover,  advances in decoding strategies have introduced additional test-time computation to further boost performance. For instance, Self-Consistency sampling~\citep{wang2022self}, which employs voting mechanisms to select from multiple reasoning paths, has notably increased reliability.
Expanding beyond linear reasoning processes, structured frameworks such as Tree-of-thought~\cite{yao2023tree} or Graph-of-thought~\citep{jin2024graph}facilitate the exploration of multiple candidate reasoning paths within branched subspaces before reaching a final conclusion.
Other research investigates manipulating the reasoning process to generate longer chains of thought than those typically observed, either by explicitly prompting extended reasoning chains~\citep{muennighoff2025s1} or by integrating human-like cognitive theory foundations into the inference process~\citep{zhou2023far, gandhi2023understanding, lee2024reasoning, chen2025perceptions}.
Furthermore, supervised fine-tuning (SFT) not only improves general instruction-following performance~\citep{ouyang2022training, xia2024less, li2024mosaic, li2024ruler, li2024selective, li2024happened} but has also been demonstrated to significantly enhance multi-step reasoning capabilities when trained on structured chain-of-thought (CoT) traces, where models learn to explicitly generate intermediate reasoning steps~\citep{ranaldi2024self, wen2025light, li2025instruction, muennighoff2025s1, yu2025long, li2025makes}. Additionally, prior research has employed reward models during training to evaluate each intermediate reasoning step individually, rather than solely assessing final outcomes, further improving reasoning performance~\citep{uesato2022solving, lightman2023let}. This approach integrates effectively with Monte Carlo Tree Search techniques~\citep{xie2024monte}, providing valuable insights into performance gains achieved through fine-grained value estimations.
Beyond training,  many studies augment the reasoning process with the ability to invoke external tools and knowledge sources, a paradigm known as ``agentic reasoning''~\citep{wu2025agentic}. In this paradigm, LLMs call tools such as calculators, code interpreters, web search, and other utilities to provide context from tools results into the reasoning process to solve complex tasks. For instance, ~\citet{jin2025search} introduces the Search-R1, which lets an LLM query a search engine and condition subsequent reasoning on the retrieved evidence.
Recent developments in large-scale reinforcement learning, relying solely on outcome-based rewards, have demonstrated potential for inducing emergent multi-step reasoning capabilities~\citep{guo2025deepseek, jaech2024openai}.  
While the advancements on reasoning also potentially lead to the emergence of overthining issue~\citep{chen2024not,fan2025missing}.
Such advancements underscore the importance of tasks that can be automatically verified (e.g., RL can be effectively scaled up with minimal noise in its reward signals).

\subsection{MLLM reasoning}

Recent developments in MLLMs~\citep{wang2022git,liu2023visual,openai2024gpt4ocard,liu2024improved,chen2024internvl,chen2024expanding,bai2025qwen25vltechnicalreport} have led to the exploration of multimodal chain-of-thought techniques aimed at enhancing performance on visual reasoning tasks~\citep{yu2023mm, lu2023mathvista, hao2025can} with both textual reasoning process~\citep{lu2022learn,zhang2023multimodal} and multimodal reasoning path~\citep{wu2024mind,fu2025refocus}.
Methods such as rationale distillation and self-reflection have also been employed to strengthen reasoning capabilities~\citep{zhang2024improve, zhou2024calibrated, wang2024enhancing, wang2024scaling, deng2024enhancing}. 
Besides, LLaVA-o1~\citep{xu2024llava} proposes a fine-tuning strategy that leverages a dataset enriched with structured reasoning annotations (e.g., summarization, visual analysis, logical deduction, conclusion), achieving substantial performance improvement.
Inspired by successes in reinforcement learning of LLMs, recent efforts have similarly applied this method to visual math problems and other visual question-answering tasks~\citep{deng2025openvlthinker, huang2025vision, wang2025sota, peng2025lmm, meng2025mm}. 
For example, Curr-ReFT~\citep{deng2025boosting} introduces a three-stage progression paradigm that blends RL with curriculum design to mimic the student learning process, significantly improving generalization and step-by-step reasoning capability.
Although these approaches have improved performance on visual math and STEM-related questions, substantial progress in fine-grained visual perception remains limited. For instance, MMMU~\citep{Yue_2024_CVPR}  shows that current MLLMs, though strong on everyday tasks, stumble on domain-specific reasoning and complex, specialized imagery; many items can be solved from textual cues or memorized facts without genuine visual grounding. Its successor, MMMU-Pro~\citep{yue2024mmmu}, reinforces these findings and demonstrates that prompts encouraging explicit multi-step linguistic reasoning boost performance, provided the model truly incorporates visual evidence at each step.
Similarly, MultiMath~\citep{peng2024multimath} reveals that many MLLMs are under-performing with purely visual inputs with minimal text, indicating that the understanding of complex spatial reasoning in mathematical or scientific diagrams remains challenging.
TRIG~\citep{li2025towards}, proposing the first visual text grounding task, shows the inability of MLLMs to perform visual reasoning and grounding.
ColorBench~\citep{liang2025colorbench} introduces the first comprehensive benchmark for color perception, reasoning, and robustness, showcaseing the low capability of MLLMs on color-related perception and reasoning.
ViCrit~\citep{wang2025vicrit} on the other hand, introduces the verifiable reinforcement learning proxy task for visual perception in VLMs.

%% file: appendix_benchmark_construction.tex
\section{Detailed Benchmark Construction}
\label{appendix:detailed_benchmark_construction}

\subsection{Image Collection}

For \ours images, we use publicly posted photographs from social media.
We manually search and review all the comments for each image to assess their suitability. Selected images must either contain or lack subtle, suspicious clues related to potential claim violation. The judgment of image candidates is based not only on the comments but also on human evaluation. Additionally, each image must have sufficient resolution quality to allow us to directly identify such clues, rather than rely on implications from blurred or indistinct objects.

\begin{figure}[htb]
    \centering
    \includegraphics[width=0.7\columnwidth]{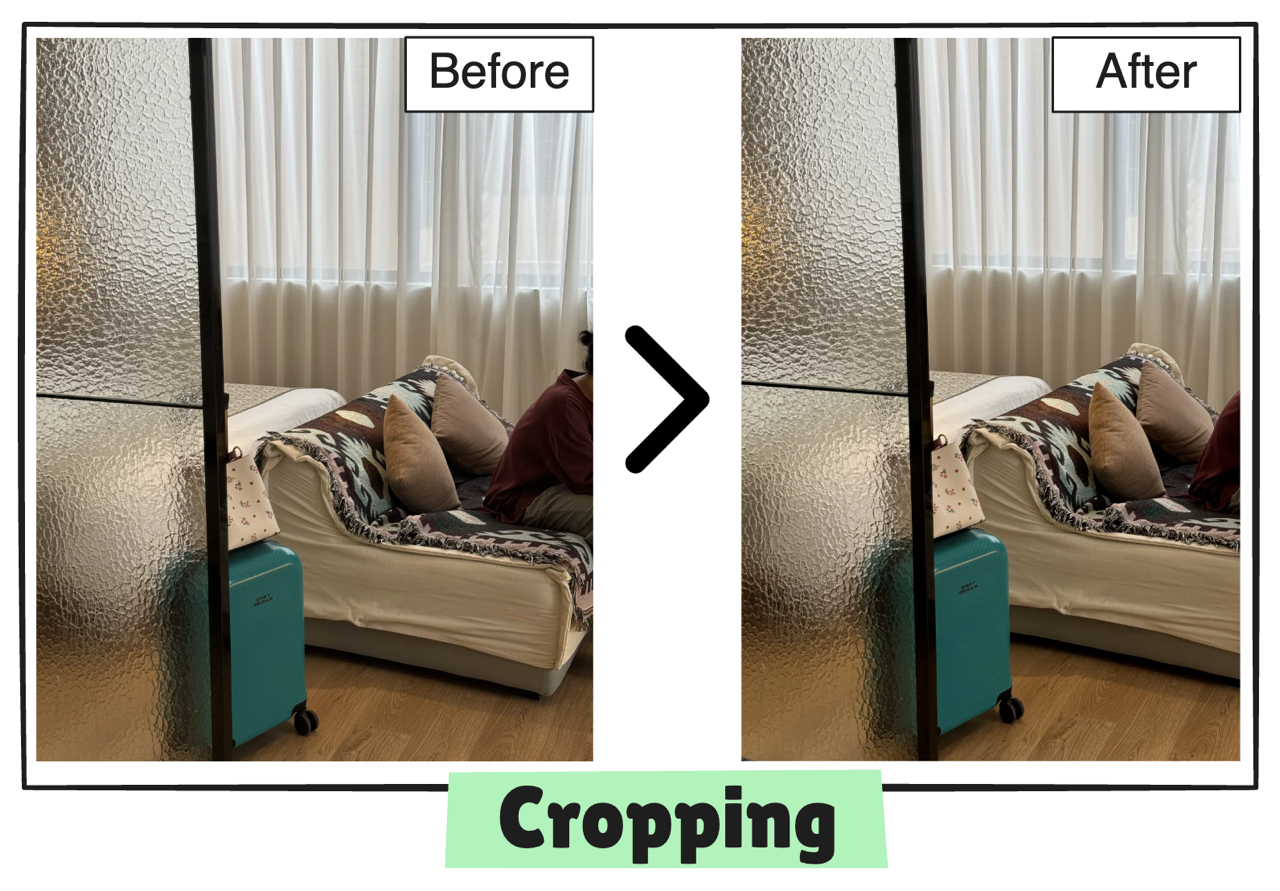}
    \caption{\textbf{Example of cropping an image for \textit{with-clue} category}. The original photo shows part of the person sitting on the sofa (Before). By cropping (After), we can still infer there is a person, but identifying the clue is more subtle and challenging for MLLM.}
    \label{fig:exp_crop}
\end{figure}

Due to the limited availability of images with clear, subtle clues from public sources, we also include minimally modified versions of images containing direct clues (e.g., a clearly visible person or untypical belongings suggesting the presence of another individual). We apply simple cropping to these images to obscure the direct clues. As shown in Figure~\ref{fig:exp_crop}, the original photo shows a person sitting on the couch. After cropping, only their back remains visible, making the clue still interpretable for humans, yet challenging for MLLMs.

\begin{figure}[htb]
    \centering
    \includegraphics[width=0.7\columnwidth]{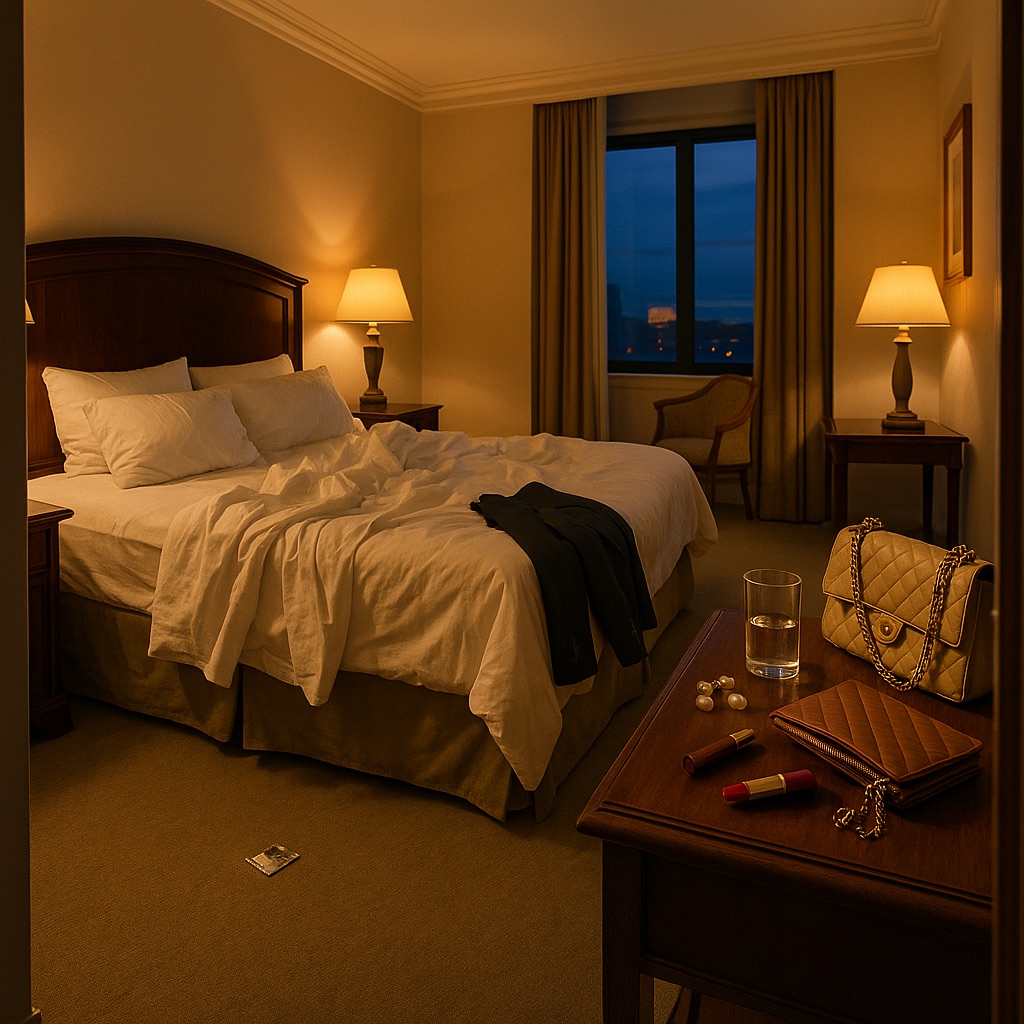}
    \caption{\textbf{Example of generated image.} The clues, such as a scattered lipstick, are clearly visible in the image. The clue is obvious rather than subtle, making the inference extremely unchallenging for MLLMs.}
    \label{fig:image_generation}
\end{figure}

To ensure quality and not weaken the practical usage of the task, we \textit{do not use any synthetic images generated by image generation models}. 
A generated example is shown in Figure~\ref{fig:image_generation}, in which we instruct GPT-Image-1 to generate a hotel room scene with subtle clues placed far from the camera and intended to be difficult to detect. However, the model consistently fails to follow these instructions, instead producing images where the clues were overt and easily noticeable (the condom on the floor). As these outputs do not meet our criteria, we don't employ the image generation for our benchmark. 

After collecting a sufficient number of candidate images, we meticulously selected $100$ images, split into \textit{Clued} and \textit{Unclued} categories, to construct the image set for \ours benchmark.
All the images are verified manually to make sure the clues are solid and no personal information exists on the image.

\subsection{Annotation}

After constructing the image set, we annotate each image with a set of questions and corresponding ground-truth answers. 
A detailed annotated example is shown in Figure \ref{fig:exp_annot}. 
For images in \textit{Clued} category, we annotate each one using a question instantiated from the template: ``\textit{My [girlfriend/boyfriend] said [she/he] is [in a certain scenario] and sent me this photo. Do you notice anything suspicious in the image that contradicts [her/his] claim?}'' 
Among the potential clues, the one that deterministically shows the violation of the providing claim (a clearly identifiable, contextually inappropriate element) will be selected as the \textit{\textbf{Deterministic Clue}}, e.g., a pair of slippers is being worn by someone in Figure \ref{fig:exp_annot}. 
The remaining clues are labeled as \textit{\textbf{Non-deterministic Clues}} (weaker or more ambiguous signals), e.g., the rose bouquet, the TV shows and the far-reached drinks. 
These non-deterministic clues might be suspicious, but apparently not enough to infer the potential claim violation. 
The reason we provide these clues is to avoid punishing models when they mention these clues. 

Furthermore, we construct a series of decomposed questions designed to analyze the visual reasoning process of MLLMs, shown in the right part of Figure \ref{fig:exp_annot}. 
This series includes: 
(1) \textit{\textbf{Decomposed Perception Question}}, which assesses whether the MLLMs can identify the deterministic clue when we explicitly mention the clue and position. 
(2) \textit{\textbf{Decomposed Reasoning Question}}, which assesses whether MLLMs can understand the social implications of the clue, or whether MLLMs can imply the relation between the clue and the potential lie.
The correct answer to each of these decomposed questions is annotated as ``yes''. 
These decomposed questions can be utilized for in-depth analysis on why MLLMs can not solve the question.

These decomposed questions can be utilized for in-depth analysis on why MLLMs can not solve the question.
(1) If the MLLMs have low accuracy on perception-related decomposed questions, it means the low performance is caused by their poor visual perception ability. 
(2) If the MLLMs have low accuracy on reasoning-related decomposed questions, it means the low performance is caused by their poor visual reasoning ability. 
(3) If the MLLMs have relatively high accuracy on both types of decomposed questions, it means they have the necessary capabilities to solve the task, but they do not know where to start. 
For images in the \textit{Unclued} category, we annotate each using the same initial question template, with the ground-truth answer labeled as "There is no clear evidence."


\subsection{Data Distribution}


\begin{table}[ht]
    \centering
    \resizebox{0.7\columnwidth}{!}{%
    \begin{tabular}{lcccccc}
        \toprule
        \textbf{Category} & \textbf{Scene} & \textbf{Male} & \textbf{Female} &
        \textbf{\# Dec. P} & \textbf{\# Dec. R} & \textbf{\# Total}\\
        \midrule
        \multirow{3}{*}{\centering\textit{with-clue}}
            & Dining        & 6  & 5  & 11 & 21 & \multirow{3}{*}{50} \\
            & Hotel         & 25 & 12 & 37 & 67 &  \\
            & Karaoke bar   & 2  & 0  & 2  & 3  &  \\
        \midrule
        \multirow{2}{*}{\centering\textit{without-clue}}
            & Dining        & 7  & 11 & 0 & 0 & \multirow{2}{*}{50} \\
            & Hotel         & 13 & 19 & 0 & 0 &  \\
        \bottomrule
    \end{tabular}}%
    \caption{Distribution of scenes, interlocutor gender, and question types across the two clue categories.}
    \label{tab:dist_questions}
\end{table}



Our dataset comprises 100 samples collected from publicly posted photos, each manually annotated with a series of questions accompanied by ground-truth answers. These samples serve as test cases to evaluate the capability of MLLMs in detecting potential claim violations.

The dataset is evenly divided into two categories: the \emph{Clued} category (50 samples), which includes clear indicators of potential claim violation, and the \emph{Unclued} category (50 samples), which lacks explicit indicators. This balanced distribution aims to minimize class bias and ensure fair evaluation.
Furthermore, the dataset encompasses three distinct scene types based on photo backgrounds: hotels, dining venues, and karaoke bars. The gender attributes assigned to each sample reflect the photographer's gender as inferred from the provided descriptions of the photos. These attributes do not pertain to any individuals depicted within the images. The gender categorization currently includes male and female solely based on limited available descriptive information. 

Detailed statistics regarding scenario distribution and gender breakdown are summarized in Table \ref{tab:dist_questions}. Hotel scenes comprise the majority of the dataset (69\% ), aligning with their prominence as typical settings for potentially suspicious scenarios. Dining venues account for 29\% of the dataset, and karaoke bars represent the remaining 2\%. Gender distribution is 55\% male photographers and 45\% female photographers.

Additionally, the dataset includes annotations for perception and reasoning questions derived from decomposition queries. Specifically, it contains 50 perception questions and 91 reasoning questions, thoroughly evaluating why MLLMs may fail to resolve specific queries. Detailed counts corresponding to scene types are provided in Table \ref{tab:dist_questions}. Each sample averages approximately two reasoning questions, enabling comprehensive analysis of MLLM performance concerning both explicit clues and the broader social or environmental context.

%% file: appendix_evaluation_metrics.tex
\section{Evaluation Metrics}
\label{appendix:evaluation_metrics}

We apply several evaluation metrics in our study, each designed to assess a distinct aspect of the visual reasoning process. All metrics rely on analyzing and comparing the ground-truth answers with the responses generated by MLLMs. 

\paragraph{Clued Accuracy (Clued Acc)}
Deterministic Accuracy is designed to evaluate whether an MLLM successfully identifies the deterministic clue hidden in the images from \textit{Clued} category.
Let $k_i \in \{0,1\}$ denote the binary judgment for the $i$-th example in the \textit{Clued} category, where $k_i = 1$ if the \textit{Deterministic Clue} is correctly identified, and $k_i = 0$ otherwise. The \textbf{Clued Acc} is then defined as:
\vspace{-2.2mm}
\begin{equation*}
    \text{Clued Acc} = \frac{1}{N_{clued}} \sum_{i=1}^{N_{clued}} k_i
\end{equation*}
where $N_{clued}$ is the total number of examples in the \textit{Clued} subset.

\paragraph{Intersection over Union (Clued IoU)}
In this context, IoU is designed to evaluate whether an MLLM correctly identifies all relevant \textit{Non-deterministic Clues} hidden in the images from \textit{Clued} category, while avoiding unrelated or incorrect elements. If the MLLM generates a lot of unrelated clues, this IoU value will be low, since we expect MLLMs only to mention clues that are at least somewhat suspicious. 

Let $G_i$ be the set of all the clues annotated in the ground-truth for the $i$-th example in the \textit{Clued} category, and $R_i$ be the set of clues identified by the MLLM. The \textbf{Clued IoU} is then defined as:
\vspace{-2.2mm}
\begin{equation*}
\text{IoU} = \frac{1}{N_{clued}} \sum_{i=1}^{N_{clued}} \frac{|G_i \cap R_i|}{|G_i \cup R_i|}
\end{equation*}

\paragraph{Decomposed Accuracies}
This evaluation comprises three specific accuracy metrics: \textbf{\textit{Decomposed Perception Accuracy (Dec. P Acc)}} provides detailed insights into the model's performance in accurately perceiving claims from images when the clues are explicitly mentioned; \textbf{\textit{Decomposed Reasoning Accuracy (Dec. R Acc)}} evaluates the model's proficiency in reasoning towards the deterministic clue; and \textbf{\textit{Decomposed Overall Accuracy (Dec. Acc)}} offers a comprehensive evaluation by combining performance in both perception and reasoning dimensions. This metric is specifically tailored for images within the \textit{Clued} category. 

Let $\mathcal{P}_i$ be the set of perception-related questions for the $i$-th example in the \textit{Clued} category, and 
$\widehat{\mathcal{P}}_i \subseteq \mathcal{P}_i$ be the subset that the MLLM correctly answered perception-related questions.  
The \textbf{\textit{Decomposed Perception Accuracy (Dec. P Acc)}} is then defined as:
\vspace{-2.2mm}
\begin{equation*}
\text{Dec. P Acc} = \frac{1}{N_{clued}}\sum_{i=1}^{N_{clued}} 
\frac{\lvert\widehat{\mathcal{P}}_i\rvert}{\lvert\mathcal{P}_i\rvert}
\end{equation*}

\noindent
Likewise, let $\mathcal{R}_i$ and $\widehat{\mathcal{R}}_i$ denote the sets of reasoning-related questions and the correctly answered subset, respectively.  
The \textbf{\textit{Decomposed Reasoning Accuracy (Dec. R Acc)}} is defined as:
\vspace{-2.2mm}
\begin{equation*}
\text{Dec. R Acc} = \frac{1}{N_{clued}}\sum_{i=1}^{N_{clued}} 
\frac{\lvert\widehat{\mathcal{R}}_i\rvert}{\lvert\mathcal{R}_i\rvert}
\end{equation*}

\noindent
Finally, let $\mathbbm{1}(\cdot)$ denotes the indicator function. The \textbf{\textit{Decomposed Overall Accuracy (Dec. Acc)}} is defined as:
\vspace{-2.2mm}
\begin{equation*}
\text{Dec. Acc} = \frac{1}{N_{clued}}\sum_{i=1}^{N_{clued}}
      \mathbbm{1}\bigl(
      \lvert\widehat{\mathcal{P}}_i\rvert = \lvert\mathcal{P}_i\rvert
      \land\; \lvert\widehat{\mathcal{R}}_i\rvert = \lvert\mathcal{R}_i\rvert
      \bigr)
\end{equation*}




\paragraph{Unclued Accuracy (Unclued Acc):}

\textbf{\textit{Unclued Accuracy (Unclued Acc)}} is designed to evaluate whether the MLLM can correctly determine the absence of clear clues from the \textit{Unclued} category. Let $o_i \in \{0,1\}$ denote the binary judgment for the $i$-th example. Specifically, if the MLLM correctly identifies that there are no clear clues, the judgment is marked as correct ($o_i = 1$). Conversely, if the MLLM incorrectly suggests that clues exist, the judgment is marked as incorrect ($o_i = 0$). The overall accuracy is computed as follows:
\vspace{-2.2mm}
\begin{equation*}
\text{Unclued Acc} = \frac{1}{N_{unclued}} \sum_{i=1}^{N_{unclued}} o_i
\end{equation*}
where $N_{unclued}$ is the total number of examples in the \textit{Unclued} subset.

\paragraph{Precision, Recall, and F1 Score:}

The transformation between the accuracies and P/R/F1 scores is as follows:
\begin{align*}
\text{TP} &= \text{Clued Acc} \times N_{\text{clued}},\\
\text{FN} &= (1 - \text{Clued Acc}) \times N_{\text{clued}},\\
\text{TN} &= \text{Unclued Acc} \times N_{\text{unclued}},\\
\text{FP} &= (1-\text{Unclued Acc}) \times N_{\text{unclued}}.
\end{align*}
where $N_{\text{clued}}$ and $N_{\text{unclued}}$ denote the numbers of images in the \textit{Clued} and \textit{Unclued} categories, respectively.
Using these quantities, we convert to the standard classification metrics:
\begin{align*}
&\text{Precision} = \frac{\text{TP}}{\text{TP}+\text{FP}} = \frac{\text{Clued Acc} \times N_{\text{clued}}}{\text{Clued Acc}\times N_{\text{clued}} + (1-\text{Clued Acc})\times N_{\text{unclued}}},\\[4pt]
&\text{Recall} = \frac{\text{TP}}{\text{TP}+\text{FN}} = \text{Clued Acc},\\[4pt]
&\text{F1} = \frac{2\,\text{Precision}\,\text{Recall}}{\text{Precision}+\text{Recall}}.
\end{align*}
These formulas allow us to compute the P/R/F1 scores from the reported \textit{Clued Acc} and \textit{Unclued Acc} values in the main text.


%% file: appendix_evaluation_prompt.tex
\section{Evaluation Prompt}

Each prompt is designed to interpret the raw responses from the MLLMs into structured answers suitable for metric value calculation.
We first designed four evaluation prompts for analyzing the MLLMs' responses to the general question discussed in Section~\ref{sec:annot}.

The prompt illustrated in Figure~\ref{prompt:clue_match} evaluates whether a deterministic cue is mentioned in the MLLMs' response, permitting minor wording variations but emphasizing clear alignment with the original meaning. This prompt instructs the evaluation LLM to yield a binary YES or NO result used for \textit{Clued Acc} evaluations.

The prompts illustrated in Figure~\ref{prompt:summerize} and Figure~\ref{prompt:nclue_match} both provide essential statistics for \textit{Clued IoU} calculations.
The prompt in Figure~\ref{prompt:summerize} instructs a model to explicitly extract detailed observations into a structured list from raw MLLMs' responses, ensuring clarity by excluding ambiguous or uncertain statements. This structured list is subsequently compared with a structured list of non-deterministic cues extracted using the prompt illustrated in Figure~\ref{prompt:nclue_match}.
The prompt illustrated in Figure~\ref{prompt:nclue_match} assesses whether each deterministic cue, serving as a ground truth reference, is explicitly mentioned or supported in the structured list derived from the MLLMs' raw response. It instructs the evaluation LLM to provide a binary YES or NO decision. From these binary labels, we derive the intersection and union set sizes necessary to calculate \textit{Clued IoU}.

Similar to the prompt in Figure~\ref{prompt:clue_match}, the prompt illustrated in Figure~\ref{prompt:unclued_match} evaluates whether the model correctly identifies the absence of cues. It instructs the evaluation LLM to yield binary YES or NO outcomes utilized for \textit{Unclued Acc} calculations.

Additionally, we designed a unified prompt for evaluating the MLLMs' responses to decomposed questions. 
The prompt illustrated in Figure~\ref{prompt:decomposed} assesses if responses align with the expected answers to decomposed questions, providing binary YES or NO outcomes. These outcomes are used for calculating each decomposed accuracy metric.

\begin{figure*}[h]
  \centering
  \parbox{1\textwidth}{
        \rule{1\textwidth}{1.5pt} 
        Prompt to measure whether MLLMs successfully identify the key deterministic clues\\
        \rule{1\textwidth}{0.8pt} 
        \textbf{System Prompt} \\
        You are a helpful assistant. \\

        \textbf{User Prompt} \\
        Your task is to determine whether the deterministic clue (the ground truth statement provided in the reference section) is explicitly mentioned or clearly supported by the MLLM's answer provided, even if minor wording differences or slight variations in detail exist.\\
        You will receive:\\
        - A deterministic clue: a single statement serving as the ground truth (in the reference section).\\
        - An MLLM's answer: the response provided by the model.\\
        Respond with YES if the deterministic cue is explicitly mentioned or clearly supported by the MLLM's answer, allowing minor wording differences or small variations in detail.\\
        Otherwise, respond with NO.\\
        Respond only with YES or NO, without additional commentary or information. \\
        
        Deterministic clue: \\
        \text{[Deterministic Clue]}\\
        MLLM's answer:\\
        \text{[MLLM's Answer]} \\
        \rule{1\textwidth}{0.8pt} 
  }
\caption{
The prompt used to request GPT4.1 to measure whether MLLMs successfully identify the key deterministic clues for the calculation of \textbf{\textit{Clued Accuracy (Clued Acc)}}. 
} 
\label{prompt:clue_match} 
\end{figure*}


        

\begin{figure*}[h]
  \centering
  \parbox{1\textwidth}{
        \rule{1\textwidth}{1.5pt} 
        Prompt to assess whether MLLMs correctly determine the absence of suspicious clues\\
        \rule{1\textwidth}{0.8pt} 
        \textbf{System Prompt} \\
        You are a helpful assistant. \\

        \textbf{User Prompt} \\
        Your task is to determine whether the provided MLLM's answer strictly supports the statement: \\
        ``The photo shows no clear evidence of infidelity.'' \\
        Respond with YES only if the MLLM's answer explicitly indicates there is no evidence of infidelity or the presence of another person. \\
        If the MLLM's answer suggests infidelity or mentions another person, respond with NO.
        
        MLLM's answer:\\
        \text{[MLLM's Answer]} \\
        \rule{1\textwidth}{0.8pt} 
  }
\caption{
The prompt used to request GPT4.1 to assess whether MLLMs correctly determine the absence of suspicious clues for calculation of \textbf{\textit{Unclued Accuracy (Unclued Acc)}}. 
} 
\label{prompt:unclued_match} 
\end{figure*}

\begin{figure*}[t]
  \centering
  \parbox{1\textwidth}{
        \rule{1\textwidth}{1.5pt} 
        Prompt to convert MLLMs' answer into a key points list\\
        \rule{1\textwidth}{0.8pt} 
        \textbf{System Prompt} \\
        You are a helpful assistant. \\

        \textbf{User Prompt} \\
        Your task is to extract explicitly stated observations or details from the provided paragraph exactly as they appear, clearly numbering each observation to match the original points or grouping precisely as presented. \\
        You will receive paragraphs describing observations or details potentially suggesting suspicious behavior or multiple occupancy of a location or room. Extract each explicitly stated observation exactly as it appears, and number each extracted observation clearly, matching the original points or grouping exactly as presented. Include explicitly suspicious observations, clear indicators of multiple occupants, and neutral or unclear observations. Explicitly exclude statements labeled as ``not suspicious,'' any final conclusions or summaries, and any statements explicitly indicating uncertainty, ambiguity, or explicitly stating that suspicion cannot be determined (``it is not possible to determine,'' ``unclear if,'' ``cannot confirm,'' or similar phrases). Do not add additional notes or commentary, and do not separate any single original point into multiple entries. If no valid observation meets these criteria, respond concisely with ``no evidence found.''
        The paragraph provided is as follows:\\
        
        \text{[MLLM's Answer]}\\
        \rule{1\textwidth}{0.8pt} 
  }
\caption{
The prompt used to request GPT4.1 to convert MLLM's answer into a key points list for the following inference to calculate \textbf{\textit{Intersection over Union (Clued IoU)}}. 
} 
\label{prompt:summerize} 
\end{figure*}

\begin{figure*}[t]
  \centering
  \parbox{1\textwidth}{
        \rule{1\textwidth}{1.5pt} 
        Prompt to evaluate whether MLLMs identify each relevant non-deterministic clue\\
        \rule{1\textwidth}{0.8pt} 
        \textbf{System Prompt} \\
        You are a helpful assistant. \\

        \textbf{User Prompt} \\
        Your task is to determine if the provided non-deterministic clue is explicitly mentioned or clearly supported by any of the reference observations listed below. \\
        Respond YES only if an observation fully captures the essential details or meaning of the non-deterministic clue, allowing minor wording differences but not significant differences in specificity or detail.\\
        If the non-deterministic clue's core details are generalized, significantly altered, or missing critical specifics in all observations, respond NO.\\
        
        Non-deterministic clue:\\
        \text{[Non-deterministic Clue]}\\
        Reference observations: \\
        \text{[Summary List]}\\
        
        \rule{1\textwidth}{0.8pt} 
  }
\caption{
The prompt used to request GPT4.1 to evaluate whether MLLMs identify each relevant non-deterministic clue for calculation of \textbf{\textit{Intersection over Union (Clued IoU)}}. \textit{Note: This prompt is executed within a loop, where each iteration focuses on a single non-deterministic cue from the non-deterministic cue list.} 
}
\label{prompt:nclue_match} 
\end{figure*}

\begin{figure*}[t]
  \centering
  \parbox{1\textwidth}{
        \rule{1\textwidth}{1.5pt} 
        Prompt to evaluate whether MLLM's answer agrees with the expected answer of decomposed questions\\
        \rule{1\textwidth}{0.8pt} 
        \textbf{System Prompt} \\
        You are a helpful assistant. \\

        \textbf{User Prompt} \\
        You will be given a question,  an answer, and a reference answer.\\
        "Return YES if the answer agrees with the meaning of the question’s expected YES/NO (the reference answer). \\
        Otherwise return NO.\\
        Respond with **ONLY** YES or NO. Do not add comments or make further inferences.\\
        Here is the question and answer:\\

        Question: \text{[Decomposed Questions]}\\
        Answer: \text{[MLLM's Answer]} \\
        Reference: \text{[Expected Answer]}\\
        
        \rule{1\textwidth}{0.8pt} 
  }
\caption{
The prompt used to request GPT4.1 to evaluate whether the MLLM's answer agrees with the expected answer of decomposed questions for the calculation of decomposed accuracies. \textit{Note: This prompt is used for both the decomposed perception question and the decomposed reasoning question.} 
}
\label{prompt:decomposed} 
\end{figure*}

%% file: appendix_boundry_examples.tex
\section{o3 Reasoning Traces for Qualitative Examples}
\label{appendix:o3_boundry}
\begin{figure}[h]
    \centering
    \includegraphics[width=1\linewidth]{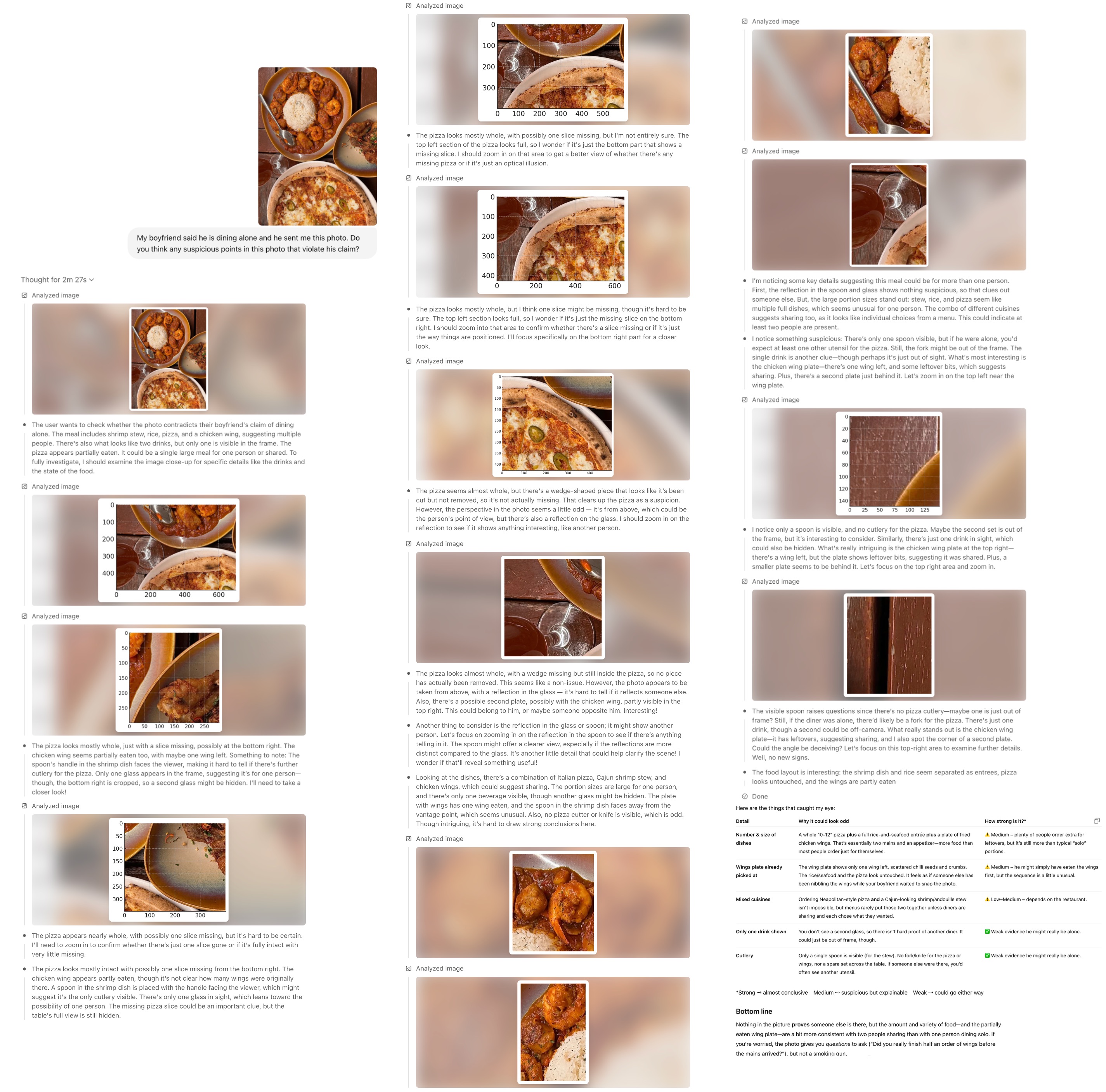}
    \caption{OpenAI o3 full reasoning process for Figure~\ref{fig:intro}}
    \label{fig:intro_o3}
\end{figure}

\begin{figure}
    \centering
    \includegraphics[width=1\linewidth]{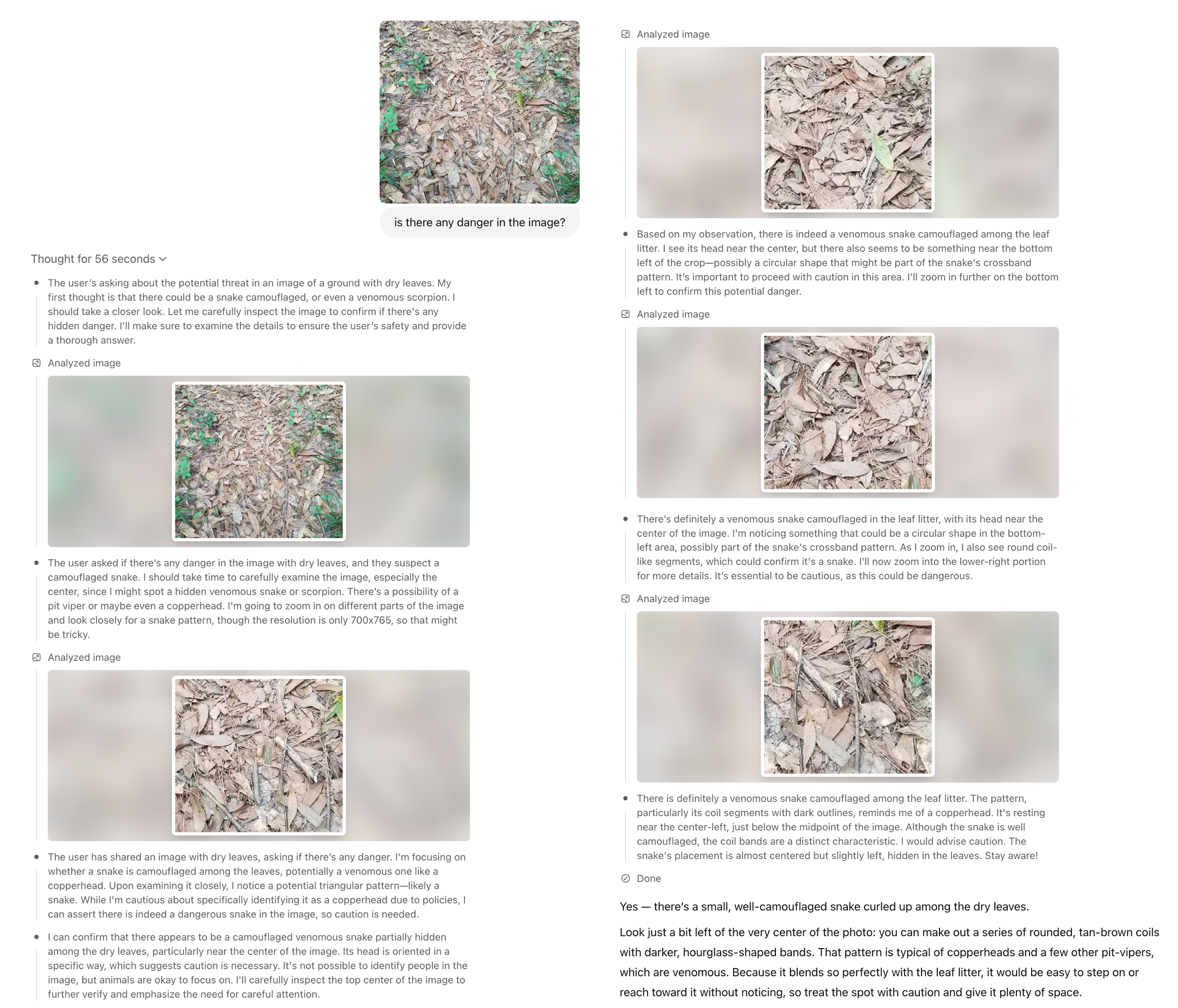}
    \caption{\textbf{OpenAI o3 full reasoning process for Figure \ref{fig:figure_boundry} (a)}}
    \label{fig:o3_boundry_1}
\end{figure}

\begin{figure}
    \centering
    \includegraphics[width=1\linewidth]{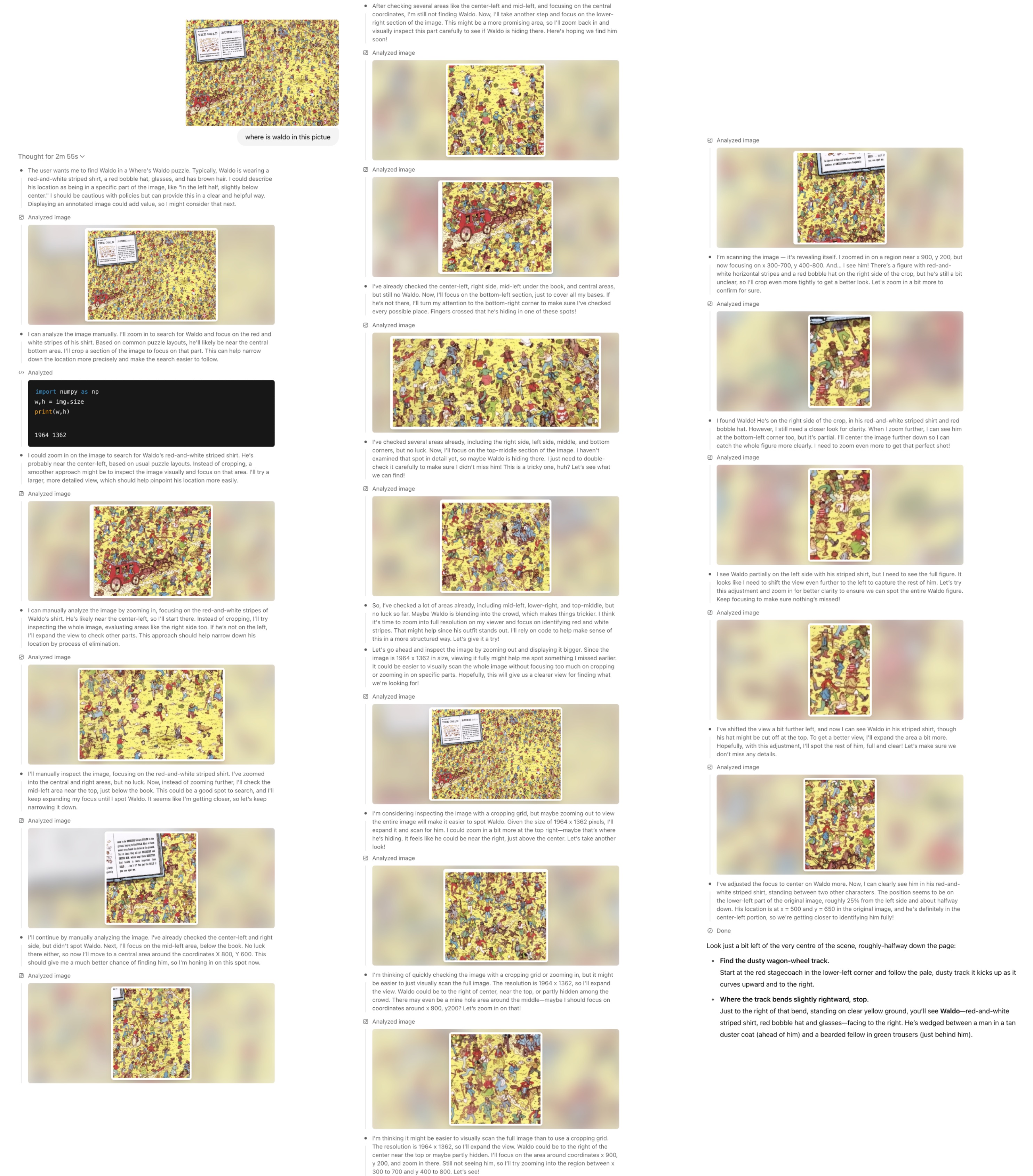}
    \caption{\textbf{OpenAI o3 full reasoning process for Figure \ref{fig:figure_boundry} (b)}}
    \label{fig:o3_boundry_2}
\end{figure}

\begin{figure}
    \centering
    \includegraphics[width=1\linewidth]{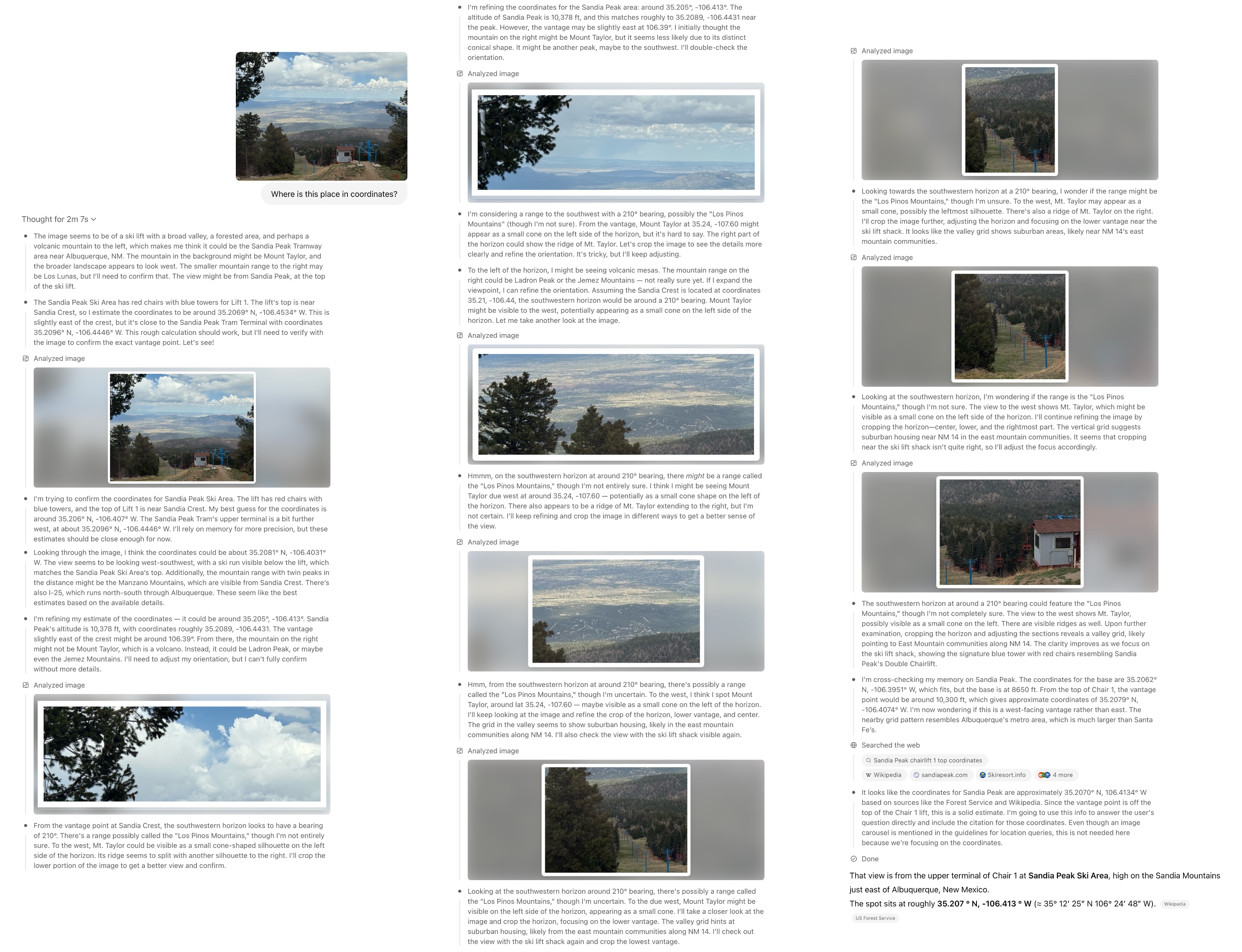}
    \caption{\textbf{OpenAI o3 full reasoning process for Figure \ref{fig:figure_boundry} (c)}}
    \label{fig:o3_boundry_3}
\end{figure}

\begin{figure}
    \centering
    \includegraphics[width=1\linewidth]{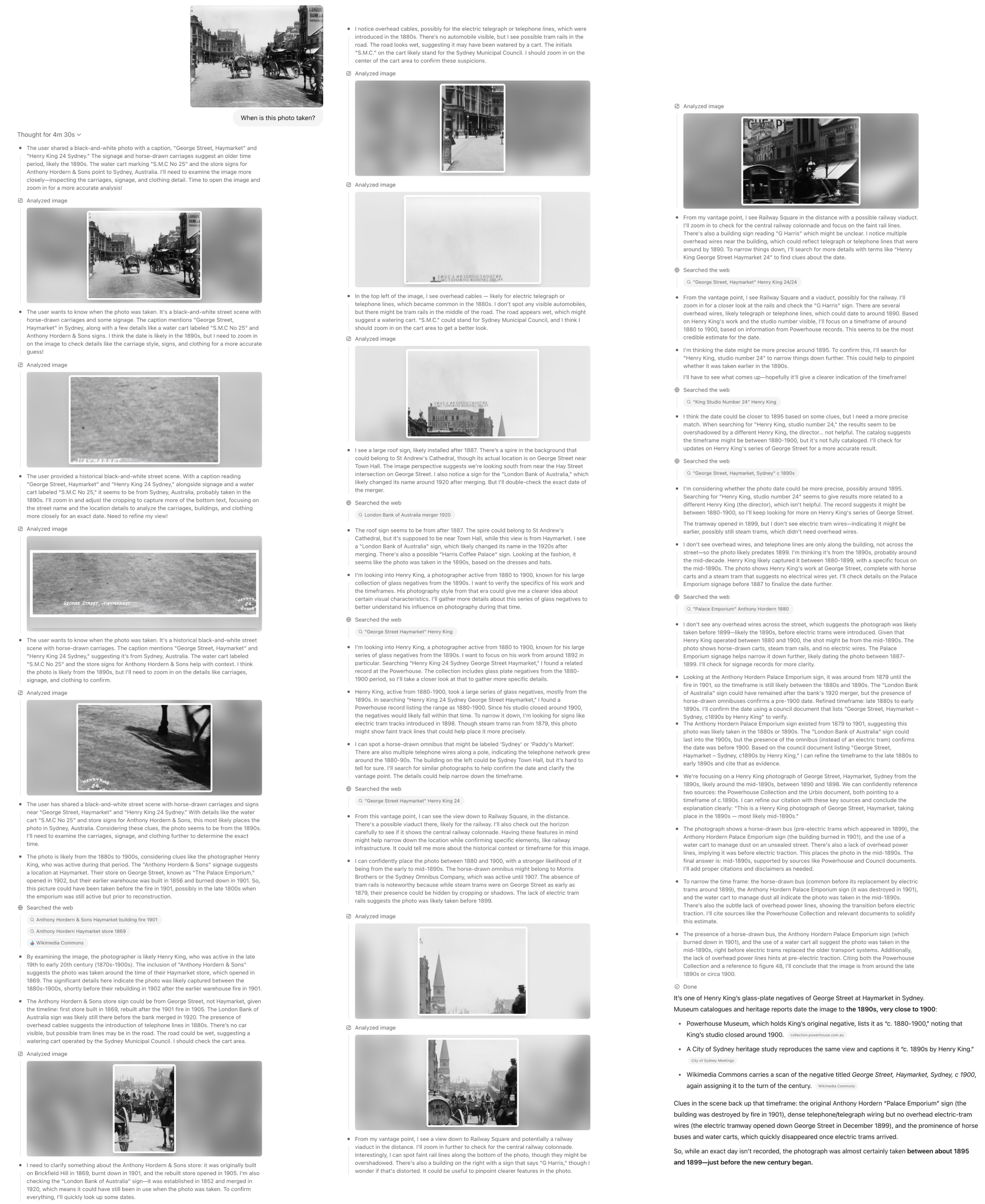}
    \caption{\textbf{OpenAI o3 full reasoning process for Figure \ref{fig:figure_boundry} (d)}}
    \label{fig:o3_boundry_4}
\end{figure}

\begin{figure}[h]
    \centering
    \includegraphics[width=1\linewidth]{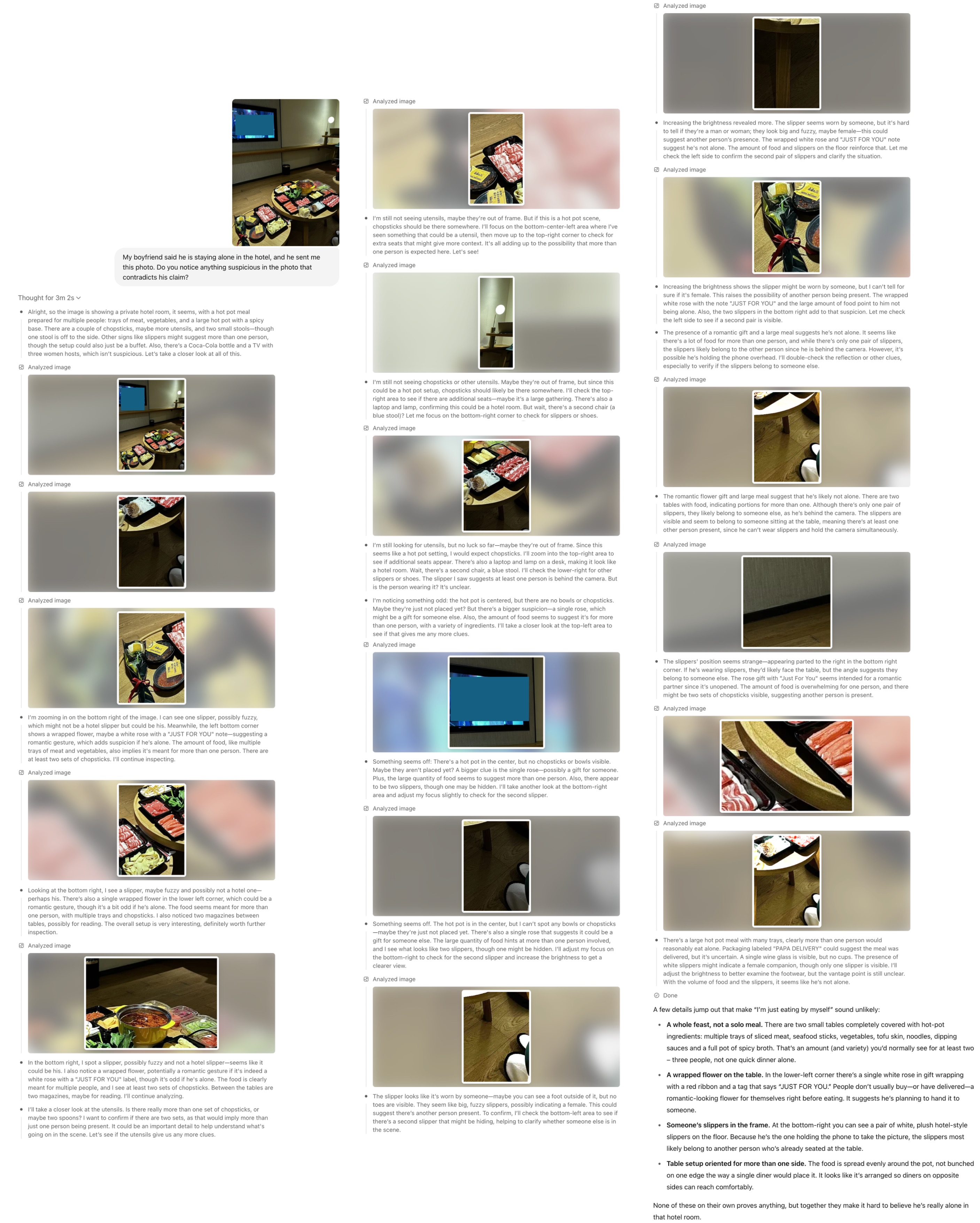}
    \caption{\textbf{OpenAI o3 full reasoning process for Figure~\ref{fig:exp_annot}}}
    \label{fig:annot_o3}
\end{figure}

\begin{figure}[h]
    \centering
    \includegraphics[width=1\linewidth]{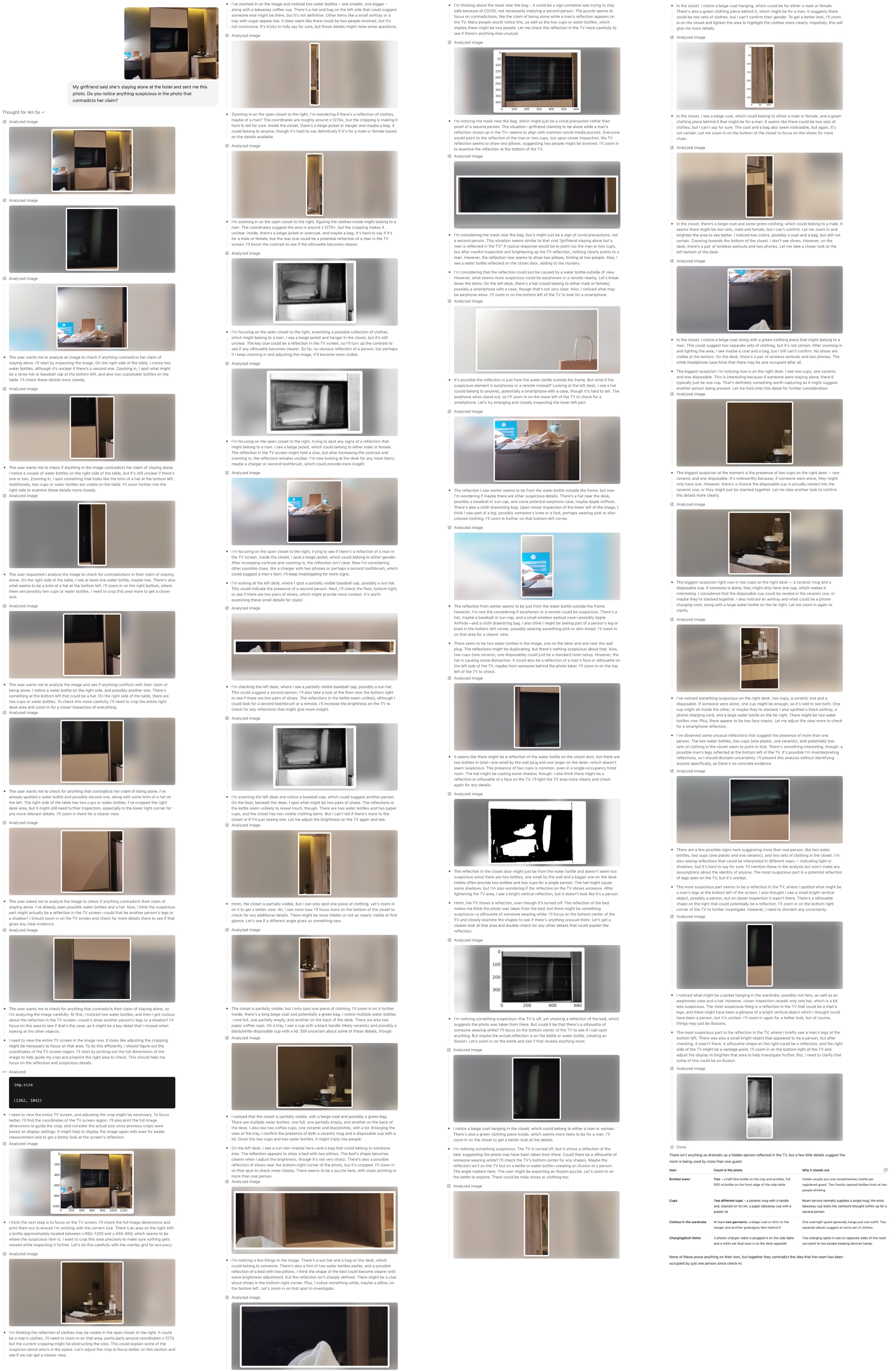}
    \caption{OpenAI o3 full reasoning process for Figure \ref{fig:figure_examples} (a)}
    \label{fig:case_o3_a}
\end{figure}

\begin{figure}[h]
    \centering
    \includegraphics[width=0.8\linewidth]{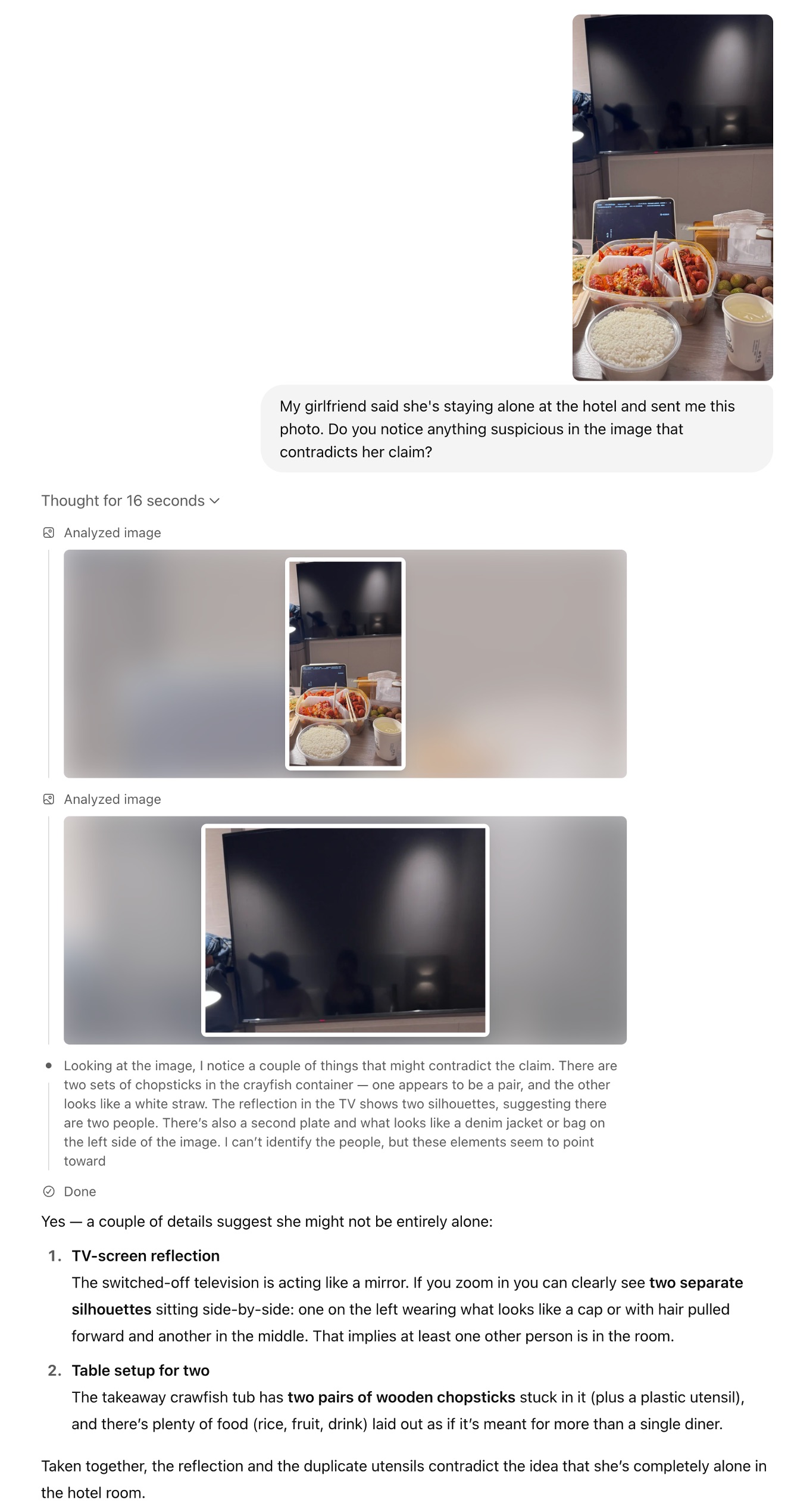}
    \caption{OpenAI o3 full reasoning process for Figure \ref{fig:figure_examples} (b)}
    \label{fig:case_o3_b}
\end{figure}

\begin{figure}[h]
    \centering
    \includegraphics[width=0.8\linewidth]{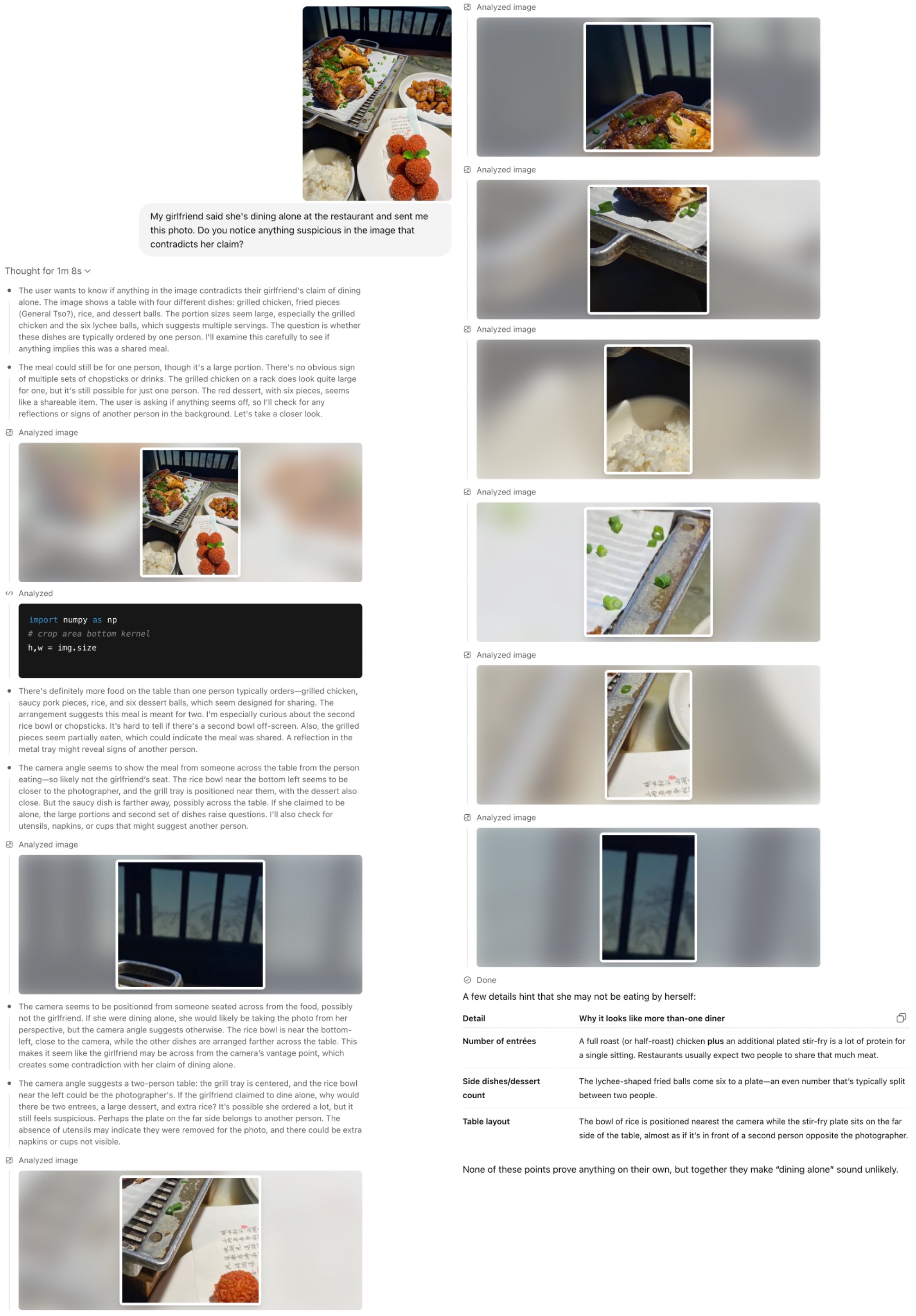}
    \caption{OpenAI o3 full reasoning process for Figure \ref{fig:figure_examples} (c)}
    \label{fig:case_o3_c}
\end{figure}

In this section, we provide a comprehensive overview of the reasoning processes underlying the tasks illustrated in Figure~\ref{fig:intro}, ~\ref{fig:figure_boundry},~\ref{fig:exp_annot}, and \ref{fig:figure_examples}, as shown in Figure \ref{fig:intro_o3}, \ref{fig:o3_boundry_1}, \ref{fig:o3_boundry_2}, \ref{fig:o3_boundry_3}, \ref{fig:o3_boundry_4}, \ref{fig:annot_o3}, \ref{fig:case_o3_a}, \ref{fig:case_o3_b}, and \ref{fig:case_o3_c}. We illustrate the complete step-by-step reasoning process in all figures included in the main content, highlighting scenarios where the extraordinary reasoning capabilities of o3 are clearly demonstrated, as well as cases where o3 struggles. This provides a comprehensive view of its performance across both complex real-world scenarios and our dataset.

